\documentclass[sigconf]{acmart}
\usepackage{graphicx}
\usepackage{amsmath}

\usepackage{amssymb}
\usepackage{multirow}
\usepackage{multicol}
\usepackage{bbm}
\AtBeginDocument{%
  }

\setcopyright{acmcopyright}

\copyrightyear{2023}
\acmYear{2023}
\setcopyright{acmlicensed}\acmConference[MM '23]{Proceedings of the 31st
ACM International Conference on Multimedia}{October 29-November 3,
2023}{Ottawa, ON, Canada}
\acmBooktitle{Proceedings of the 31st ACM International Conference on
Multimedia (MM '23), October 29-November 3, 2023, Ottawa, ON, Canada}
\acmPrice{15.00}
\acmDOI{10.1145/3581783.3611959}
\acmISBN{979-8-4007-0108-5/23/10}

\begin{document}

\title{Rethinking the Localization in Weakly Supervised Object Localization}


\author{Rui Xu}
\affiliation{%
  \institution{School of Computer Science, \\ Wuhan University,}
  \city{Wuhan}
  \country{China}}
\email{rui.xu@whu.edu.cn}

\author{Yong Luo}
\authornote{Corresponding authors: Yong Luo and Bo Du.}
\affiliation{%
  \institution{Wuhan University \& \\ Hubei Luojia Laboratory,}
  \city{Wuhan}
  \country{China}}
\email{luoyong@whu.edu.cn}

\author{Han Hu}
\affiliation{%
  \institution{Beijing Institute of Technology,}
  \city{Beijing}
  \country{China}}
\email{hhu@bit.edu.cn}

\author{Bo Du}
\authornotemark[1]
\affiliation{%
  \institution{Wuhan University \& \\ Hubei Luojia Laboratory,}
  \city{Wuhan}
  \country{China}}
\email{dubo@whu.edu.cn}

\author{Jialie Shen}
\affiliation{%
  \institution{City, University of London,}
  \city{London}
  \country{United Kingdom}}
\email{jialie@gmail.com}

\author{Yonggang Wen}
\affiliation{%
  \institution{Nanyang Technological University,}
  \city{Singapore}
  \country{Singapore}}
\email{ygwen@ntu.edu.sg}


\renewcommand{\shortauthors}{Xu et al.}

\begin{abstract}
  Weakly supervised object localization (WSOL) is one of the most popular and challenging tasks in computer vision. This task is to localize the objects in the images given only the image-level supervision. Recently, dividing WSOL into two parts (class-agnostic object localization and object classification) has become the state-of-the-art pipeline for this task. However, existing solutions under this pipeline usually suffer from the following drawbacks: 1) they are not flexible since they can only localize one object for each image due to the adopted single-class regression (SCR) for localization; 2) the generated pseudo bounding boxes may be noisy, but the negative impact of such noise is not well addressed. To remedy these drawbacks, we first propose to replace SCR with a binary-class detector (BCD) for localizing multiple objects, where the detector is trained by discriminating the foreground and background. Then we design a weighted entropy (WE) loss using the unlabeled data to reduce the negative impact of noisy bounding boxes. Extensive experiments on the popular CUB-200-2011 and ImageNet-1K datasets demonstrate the effectiveness of our method.
\end{abstract}




\maketitle

\section{Introduction}
\label{sec:intro}

The development of deep neural networks has greatly advanced a wide range of multimedia applications.
Yet the training of deep learning models usually requires a large amount of accurately annotated data. The collection of labeled data could be extremely labor intensive and time consuming, especially when fine-grained annotation is required, e.g., the object-level labels for localization and pixel-level labels for segmentation. In order to mitigate this issue, there is an increasing interest on designing weakly supervised approaches.
In this paper, we mainly focus on the task of weakly supervised object localization (WSOL), which aims to localize the objects given the image-level labels. It is one of the most fundamental and challenging tasks in computer vision, and can facilitate many multimedia applications.

\begin{figure}[!t]
  \centering
  \includegraphics[width=3.5in]{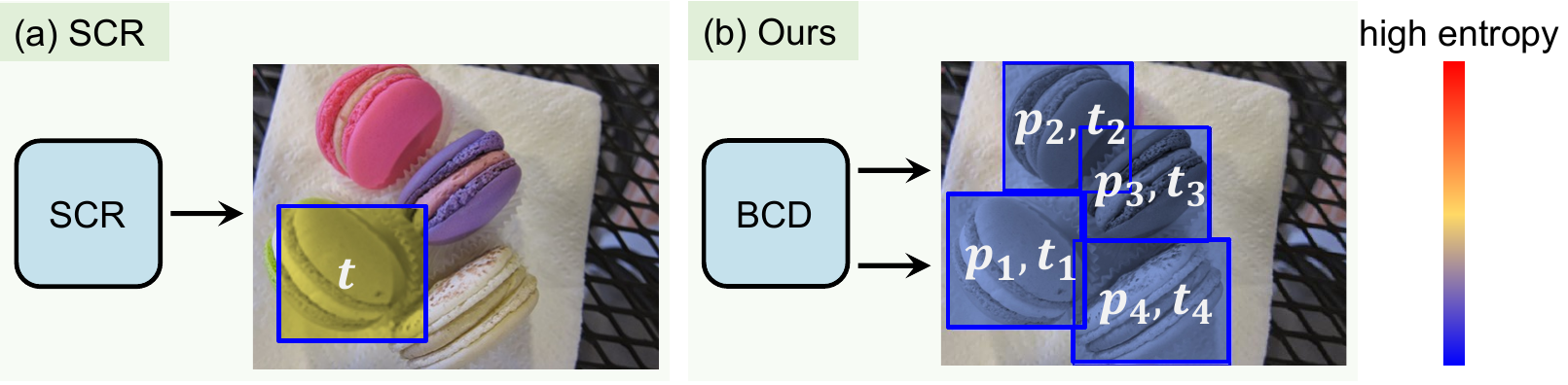}
  \caption{(a) The localization method based on SCR in current separated localization-classification WSOL solution, which provides only one output bounding box; (b) Our BCD predicts multiple bounding boxes together with highly confident foreground probabilities; \textbf{\textit{t}} denotes the predicted locations of the objects and \textbf{\textit{p}} denotes the predicted foreground probabilities.}
  \label{fig:PSOLvsOurs}
  \vspace{-0.1in}
\end{figure}

There have been dozens of WSOL approaches during the past decades. Recently, the separated localization-classification pipeline \cite{GC-Net20, PSOL20, SLTNet21, C2AM22}, which regards WSOL as two separate sub-tasks: class-agnostic object localization and object classification, achieves the state-of-the-art (SOTA) performance, and results in paradigm shift from the previous 
unified localization-classification methods \cite{CAM16, ADL19, FAM21, I2C20, SPA21, ORNet21} to the separated ones for WSOL.
The object classification is performed using the off-the-shelf classification models. As for the class-agnostic object localization, pseudo ground-truth bounding boxes are usually first generated \cite{PSOL20, SLTNet21, C2AM22}. For example, in PSOL \cite{PSOL20}, the deep descriptor transformation (DDT) \cite{DDT19} is directly adopted to generate the bounding boxes. The SLT-Net \cite{SLTNet21} dynamically improves the accuracy of the pseudo bounding boxes by strengthening the learning tolerance to the semantic errors and image transformations. The C$^{2}$AM \cite{C2AM22} proposes cross-image foreground-background contrastive learning to generate accurate pseudo bounding boxes without any supervision.
Then a single-class regression (SCR) \cite{SCR14} is trained using these generated bounding boxes.

Despite the SOTA performance of the separated localization-classification pipeline, a major drawback is the adopted simple SCR \cite{SCR14} for localization. SCR \cite{SCR14} is only able to provide one output bounding box for an image, and thus inadequate and inflexible when coping with images containing multiple objects, which are common in the real-world applications. 
In addition, the generated pseudo bounding boxes may be inaccurate or even completely wrong, but such noisy labels are directly utilized for training without any careful consideration.

To overcome these drawbacks, we propose a novel WSOL method termed \textit{Weighted ENtropy guided binary-class Detector} (WEND). In particular, we first propose to replace SCR \cite{SCR14} with a binary-class detector (BCD), which can naturally output multiple bounding boxes, and be trained in a binary classification manner by discriminating the foreground and background. This can improve both the flexibility and accuracy since any competitive detectors could be incorporated.
Then to alleviate the negative impacts of the noisy bounding boxes on the training of detector, we further propose a weighted entropy (WE) loss by making use of the large amounts of unlabeled data.
The entropy minimization is able to reduce the uncertainty when discriminating the foreground and background. Considering that the background parts are usually much more than the foreground objects, we re-weight the entropy loss to down-weight the overly confident background and make the detector pay more attention on the less confident foreground.
As shown in Figure~\ref{fig:PSOLvsOurs}, compared with the localization approach adopted in the current separated localization-classification pipeline, our WEND is able to predict multiple bounding boxes, and the predictions are of low entropy, i.e., high confidence.


Our contributions can be summarized as follows:

\begin{itemize}
    \item We rethink and analyze the localization in the SOTA separated localization-classification pipeline for WSOL, and propose a novel method to enhance the localization performance.
    \item We propose to utilize a detector trained in a binary-classification manner to localize multiple objects. The detector is more flexible and reliable than the single-class regression, which can only localize one object.
    \item We design a weighted entropy loss as an additional unsupervised constraint to reduce the negative effect of noisy pseudo bounding boxes and deal with the foreground-background imbalance in the training of the binary-class detector.
\end{itemize}
Extensive experiments are conducted on two widely used WSOL datasets: CUB-200-2011 \cite{WahCUB_200_2011} and ImageNet-1K \cite{ILSVRC15}. The results demonstrate that our method significantly outperforms the SOTA WSOL approaches.

\section{Related Works}
\label{sec:rw}

In weakly supervised object localization (WSOL), the localization of objects is usually conducted under the supervision of image-level labels.
A popular solution for WSOL is to utilize CAM \cite{CAM16}, which generates class activation maps by aggregating deep feature maps using the final fully connected layer of the classifier. However, the classifier mainly focuses on the discriminative parts of the objects, and this may lead to inaccurate localization.
To tackle this issue, many approaches \cite{ADL19,FAM21,I2C20,SPA21} are proposed by, such as, hiding the most discriminative part and highlighting the informative region of deep feature maps~\cite{ADL19}.
While these approaches mostly utilize the deep feature maps, the low level features are leveraged in \cite{ORNet21} for class activation map generation.
Some other works conducted WSOL under weaker supervision.
For example, SCDA \cite{SCDA17}, MO \cite{MO19}, DDT \cite{DDT19}, and PSY \cite{PSY20} just utilize the models pre-trained on ImageNet \cite{ILSVRC15}.
In \cite{JGP22}, the authors design a siamese network to encode the transformed image pairs, and propose a joint graph partition mechanism to enhance the object co-occurrent regions. There also exist some completely unsupervised approaches \cite{UODL15,DUSD18,ReDraw19,DiLo21} or some weakly-supervised approaches without utilizing pre-trained weights, but the performance are usually unsatisfactory.

Different from the approaches that unifies the classification and localization in a single framework, several recent works divide WSOL into two sub-parts: object classification and class-agnostic object localization \cite{GC-Net20, PSOL20, SLTNet21, C2AM22}, where some ready-made classifiers are directly employed for classification. In regard to the localization, GC-Net \cite{GC-Net20} trains a regression using the geometry constraint and a pre-trained classifier to distinguish the obtained foreground and background. PSOL \cite{PSOL20} directly generates pseudo ground-truth bounding boxes using DDT \cite{DDT19} for training. In \cite{SLTNet21}, the authors propose to strengthen the network's tolerance to the classification errors and image transformations for consistently optimizing the generated pseudo bounding boxes. While the image-level labels are involved in these approaches, C$^{2}$AM \cite{C2AM22} uses unsupervised contrastive learning to discriminate the cross-image foreground objects and backgrounds for pseudo bounding box generation.
These approaches have been demonstrated to usually achieve very competitive performance, and the separated localization-classification pipeline has become a new paradigm.

Unlike the regressor commonly used in WSOL tasks which merely and directly regresses the four coordinates of the object location, the object detector regresses the location of the object from multiple references that are close to the object. In this work, we regard modern detectors as two types, NMS-based detectors and NMS-free detectors. Their main difference exists in that during training how they match the predictions with the ground truths. NMS-based detectors \cite{tip20/foveabox, cvpr20/atss, FCOS19, eccv18/cornernet, focalloss17, fasterrcnn15, miccai22/lssanet} first generate a large number of predetermined references (anchors or points) and label each of them as positive or negative based on its overlap with the ground truths. This kind of mechanism cannot prevent an object from being predicted by multiple references, leading to duplicates and requiring complex post-processing. NMS-free detectors \cite{iclr22/sparseDETR, DeformableDETR21, iccv21/dynamicDETR, cvpr21/sparsercnn, eccv20/detr} utilize the attention mechanism to capture the pairwise relations between the references, and perform one-to-one matching between them and the ground truths, thus avoiding duplicates.

\begin{figure*}[!t]
  \centering
  \includegraphics[width=6.5in]{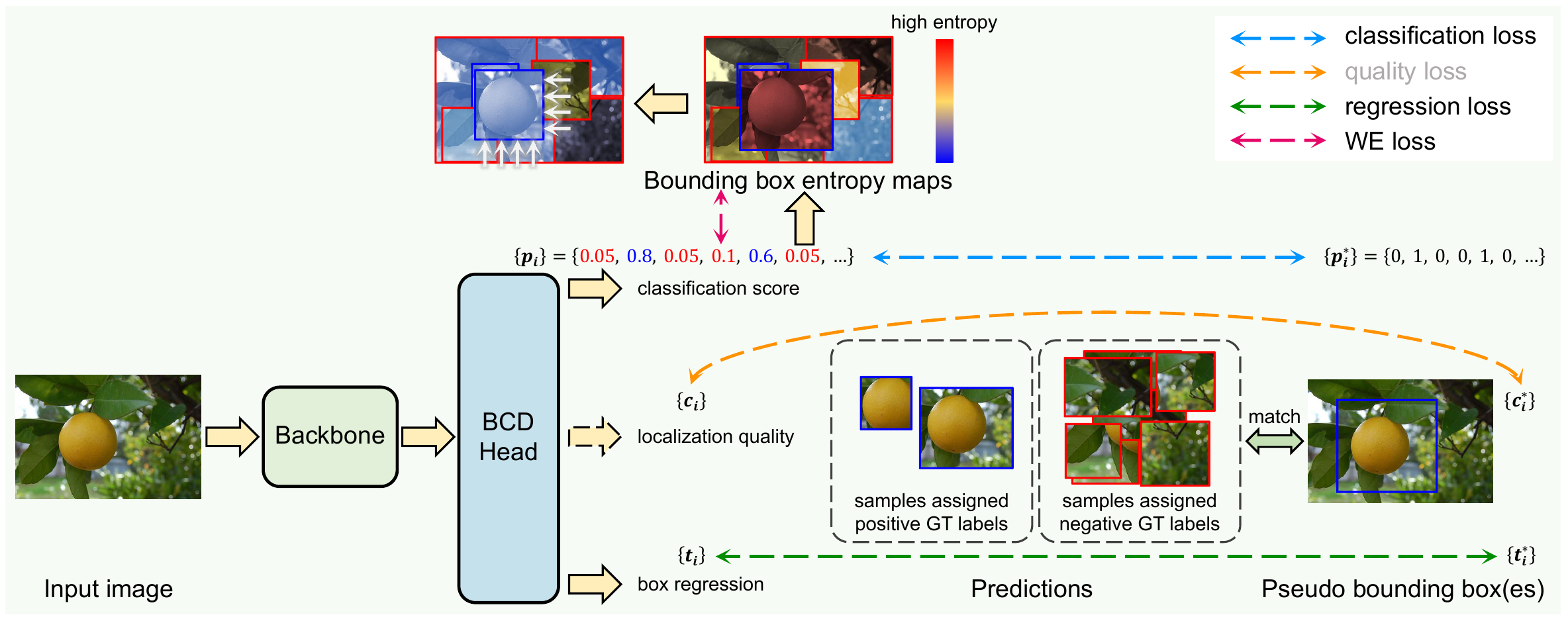}
  \caption{Overview of our \textit{Weighted ENtropy guided binary-class Detector} (WEND) for localization. In the training phase, a multi-output detector (such as RPN+R-CNN) is trained in a binary-classification manner by discriminating the foreground and the background, where the class-agnostic pseudo bounding boxes generated by existing approaches (such as C$^2$AM) are utilized as the ground-truth labels. The output positive (blue) and negative (red) probabilities $\{p_i\}$ and locations $\{t_i\}$, (as well as the localization quality $\{c_i\}$ if possible) are compared with the pseudo bounding boxes for training. In addition, an unsupervised weighted entropy (WE) constraint is applied to the classifier of the binary-class detector (BCD) for dealing with the negative impact of the noisy pseudo labels. By enforcing the weighted entropy minimization on the predicted probabilities, the foregrounds can be identified with higher confidence, and the bounding boxes can be further refined.
  }
  \label{fig:ours}
\end{figure*}

\section{Methodology}
\label{sec:method}

In this section, we analyze the localization in the separated localization-classification framework \cite{GC-Net20, PSOL20, SLTNet21, C2AM22} for weakly supervised object localization (WSOL). Then we introduce our proposed \textit{Weighted ENtropy guided binary-class Detector} (WEND) to improve the localization performance. 

\subsection{Analysis}
The highly effective separated localization-classification pipeline treats WSOL as two sub-tasks: object classification and class-agnostic object localization. For the former sub-task, existing classification model is directly adopted. In regard to the latter localization sub-task, pseudo ground-truth bounding boxes are first generated.
Afterwards, the single-class regression (SCR) \cite{SCR14} is performed using the training images and the corresponding generated pseudo bounding boxes.

Although this separated localization-classification pipeline achieves the SOTA performance in WSOL, it is limited in that: 
\begin{itemize}
    \item The simple localization model SCR \cite{SCR14} can only provide one bounding box output per image. For datasets such as the ImageNet-1K \cite{ILSVRC15}, some images may have more than one ground-truth bounding box. Therefore, the single output may be inappropriate.
    \item It is inevitable that some generated pseudo ground-truth bounding boxes, e.g., by utilizing either DDT \cite{DDT19} or C$^{2}$AM \cite{C2AM22}, are not optimal or even completely wrong, while there is no design in the simple SCR \cite{SCR14} to deal with such noisy labels.
\end{itemize}
Therefore, there is still much room for improvement in the separated localization-classification pipeline.

\subsection{The Proposed Method}
To deal with the aforementioned two drawbacks, we propose the \textit{Weighted ENtropy guided binary-class Detector} (WEND).
Following the separated localization-classification pipeline, our WEND consists of two separate branches, where the object classification branch directly adopts an existing model.
Our class-agnostic object localization branch is illustrated in Figure~\ref{fig:ours}, where we first adopt the high-performing
C$^{2}$AM \cite{C2AM22} to generate pseudo bounding boxes.
Then for the localization, we propose to employ a binary-class detector (BCD) and design an auxiliary unsupervised weighted entropy (WE) loss.
The BCD, as a replacement of the SCR, is able to output multiple localization boxes, and trained by discriminating the foreground objects from the backgrounds.
Any existing detector can be incorporated in BCD. 
In order to handle the noisy labels, we propose a WE loss to reduce the uncertainty in the prediction, and the imbalance between the foregrounds and backgrounds are considered by assigning different weights. More details are depicted as follows.

\subsubsection{Binary-Class Detector}

After generating pseudo bounding boxes, the object localization sub-task becomes a pseudo supervised problem, where the separated localization-classification pipeline performs SCR \cite{SCR14} on these pseudo bounding boxes and suffers the above-mentioned weaknesses. Here we address the problem of inflexible outputs by simply introducing a detector, which can be any modern object detectors. 
The key is how to train such detector given the pseudo bounding boxes.

In order to realize the class-agnostic localization, we regard the generated bounding boxes as pseudo ground-truth boxes and assign the same category of the foregrounds to them. In this way, the localization can be realized through the standard detection steps. The feature map of the input image is first extracted using a conventional CNN backbone, which can be the VGG \cite{vgg15}, ResNet \cite{resnet16}, Inception \cite{inception16}, or GoogLeNet \cite{googlenet15}. Then in our work, the detection head is directly applied to the feature map for generating the predictions, which are comprised of the predicted classification probabilities (as well as the localization qualities if possible) and the predicted bounding boxes. It should be noted that here we treat the encoder-decoder transformer \cite{transformer17} in the transformer-based detectors as a part of the detection head. Some configurations of the detection head are adjusted in terms of the number of classes, the design of anchors or the number of queries according to our kind of ground-truth assignments and the characteristics of the WSOL task, where most of the images are object-centric. 
Then during the training process, the pseudo ground-truth bounding boxes are compared with the predictions of the detector through Intersection-over-Union (IoU) overlap or one-to-one matching for assigning the labels. For instance, for the anchor-based detectors, an anchor is assigned a positive label if it has an IoU overlap equal to or higher than a predetermined foreground IoU threshold with any pseudo ground-truth boxes. If there are no positive anchors that meet this criterion, the anchor with the highest IoU overlap with a pseudo ground-truth box is considered as positive. On the flipside, an anchor is assigned a negative label if is has an IoU less than a predetermined background IoU threshold with all the pseudo ground-truth boxes. All the other anchors are ignored and do not contribute to the training process. The matching results are utilized to compute the pseudo supervised loss, which consists of the classification loss and the regression loss:

\begin{align}
\nonumber
  L_{sup} = L(\{p_{i}\}, \{t_{i}\}) = \lambda_{1}\frac{1}{N_{cls}}\sum_{i}L_{cls}(p_{i}, p_{i}^{*}) \\
  +\lambda_{2}\frac{1}{N_{reg}}\sum_{i}p_{i}^{*}L_{reg}(t_{i}, t_{i}^{*}),
  \label{sup_loss}
\end{align}
where $i$ is the index of a sample in a mini-batch. This mini-batch is a random selection of the training samples, and the ratio of positive to negative samples are also specified manually. $p_{i}$ is the predicted class probability of the $i$-th sample being a foreground object. $p_{i}^{*}$ is the ground-truth label of the $i$-th sample; it is 1 if the sample is positive, and is 0 if the sample is negative. $t_{i}$ and $t_{i}^{*}$ are vectors denoting the 4 parameterized coordinates of the predicted bounding box and of the corresponding pseudo ground-truth box, respectively. The $L_{cls}$, which can be the binary cross-entropy loss, measures the classification error between the assigned foreground/background labels and the predicted class probabilities, and the $L_{reg}$, which can be the smooth $L_{1}$ loss or GIoU loss, measures the localization error between the pseudo ground-truth boxes and the predicted boxes. These two terms are then normalized by the numbers of samples $N_{cls}$ and $N_{reg}$ separately, and weighted by the balancing parameter $\lambda_{1}$ and $\lambda_{2}$. Some detectors also utilize an auxiliary branch to predict the localization quality, e.g. the overlap of the predicted bounding box and the ground-truth box, which is usually denoted by IoU, or the deviation between the center points of the predicted bounding box and the ground-truth box, which is often denoted by centerness. This branch is trained using the binary cross-entropy loss, which measures the difference between the predicted localization quality and the ground truth.

\begin{table}[!t]
\small
    \centering
    \caption{Comparison of the parameters and running speeds.} 
    \setlength{\tabcolsep}{6mm}{
    \begin{tabular}{ccc} \toprule
    Method          & \#Params & runtime (s/batch) \\ \midrule
    SCR             & 24.55M   & 0.331             \\
    200-class FCOS  & 102.79M  & 0.382             \\
    1000-class FCOS & 117.54M  & 0.408             \\
    BCD (Ours)      & 99.12M   & 0.245             \\ \bottomrule
    \end{tabular}}
    \label{tab:compute}
\end{table}

Since the detector naturally gives multiple bounding box outputs, the drawback of inflexible outputs is easily removed. 
It is noteworthy that the previous works suggest that detection models are too heavy for the object localization task. On the one hand, the calculation burden of the current detectors themselves is heavy. On the other hand, datasets like ImageNet \cite{ILSVRC15} in the object localization task have much more classes than the ones in the object detection task, e.g., 1000 vs. 200. But our work subtly adjusts the number of samples for training in the detector, and proposes to only classify the foreground objects and the backgrounds, thus reducing the computational burden. In this way, the detection models can be applied in the large-scale localization field. 
To illustrate our computation reduction quantitatively, we conduct the model's parameter and speed analysis of the SCR, the detectors that could be used for the WSOL tasks, and our BCD. The results are listed in Table~\ref{tab:compute}. The results of the 200-class FCOS and 1000-class FCOS show that introducing an additional detector increases the runtime. However, by using our operations, our BCD even runs faster than the SCR. 

\subsubsection{Weighted Entropy Loss}

In order to reduce the negative impact of the noisy pseudo bounding boxes, we propose a WE loss as an additional unsupervised loss. 
Entropy minimization is a popular strategy for training the unlabeled data in the semi-supervised and unsupervised image classification tasks \cite{semi04,mixmatch19,fs20,tent21,shot22}. It encourages the classifier to divide the unlabeled data into well separated categories. In this work, we utilize the entropy loss \cite{entropy1948} as an additional constraint for the unlabeled data's foreground-background classification. This constraint is independent of the fixed generated pseudo bounding boxes, and is only related to the predictions. Besides, we assume that due to the weakly localization ability of the classifier, the entropy loss \cite{entropy1948} is also beneficial for dealing with the offsets of the boxes. However, the original entropy minimization often assumes a uniform distribution of the categories, while there exists a foreground-background imbalance problem in the object localization using a detector \cite{od_imbalance21}. The predictions of the background are dominant and overly confident, whereas the predictions of the foreground are the opposite. Simply adopting the entropy loss \cite{entropy1948} can bias the detector's predictions even further. Inspired by the success of Focal loss \cite{focalloss17}, we come up with an elegant solution to deal with this foreground-background imbalance issue in localization. Namely, we design an extra WE loss aiming at down-weighting the easy samples and focusing on the hard ones for the classifier of the detector:


\begin{align}
  L_{unsup} = L(\{p_{i}\}) = \frac{1}{N_{cls}}\sum_{i}H_{cls}(p_{i}, \tau_{1}, \tau_{2}), \\
  \nonumber
  H_{cls}(p_{i}, \tau_{1}, \tau_{2}) = - (\mathbbm{1}(p_{i} < \tau_{1})(1 - \alpha)p_{i}^{\gamma} \\ + \mathbbm{1}(p_{i} > \tau_{2})\alpha(1 - p_{i})^{\gamma})p_{i}\log(p_{i}),
  \label{unsup_loss}
\end{align}
where $\tau_{1}$ and $\tau_{2}$ are the confidence thresholds for the choice of the modulating factor $p_{i}^{\gamma}$ or $(1 - p_{i})^{\gamma}$, 
which is also solely related to the predicted probability rather than the generated pseudo bounding box labels. The $\tau_{1}$ and $\tau_{2}$ ensure the modulating factor being proper from two aspects. First, they help to determine the well-classified samples. Second, at the beginning of the detector training, many samples which have predicted class probabilities between $\tau_{1}$ and $\tau_{2}$ are neglected, and not involved in calculating the WE loss. This alleviates the perturbation of the unsupervised WE loss on the initial training and avoids the collapse of training. As the training progresses, more and more samples are used with appropriate modulating factor. It is noteworthy that our proposed modulating factor choice mechanism here actually follow the strategy of self-paced learning \cite{ijcai15/SPL,ijcai21/SAIL}.
Then the selected modulating factor is able to down-weight the well-classified samples' entropy and focus on the hard ones. As in \cite{focalloss17}, $\gamma$ is a tunable focusing parameter utilized to control the degree of down-weighting, and $\alpha$ is another hyperparameter for dealing with the class imbalance. 

To summarize, given the pseudo ground-truth bounding boxes generated by the effective C$^{2}$AM \cite{C2AM22} and the training images, we train a BCD for the binary class-agnostic object localization sub-task by minimizing the following loss:
\begin{equation}
  L_{loc} = \eta L_{sup} + L_{unsup},
  \label{total_loss}
\end{equation}
where $\eta \geq 0$ is a balancing hyper-parameter.

\section{Experiments}

\subsection{Experimental Setups}

\textbf{Datasets.} We evaluate our method on two widely used WSOL datasets: the CUB-200-2011 \cite{WahCUB_200_2011} and ImageNet-1K \cite{ILSVRC15}. The CUB-200-2011 \cite{WahCUB_200_2011} is a challenging dataset of 200 bird species, which is comprised of 5,994 images for training and 5,794 images for test. Each image in the CUB-200-2011 \cite{WahCUB_200_2011} is annotated with a bounding box and a class. The ImageNet-1K \cite{ILSVRC15} is a large-scale visual recognition dataset of 1000 categories, which consists of 1,281,167 images for training and 50,000 images for validation. Each image in the ImageNet-1K \cite{ILSVRC15} is assigned one class, and each validation image is labeled one or more bounding boxes. In the setting of WSOL, only the image-level supervision, namely the class label, is allowed to be used during training. In our work, it is utilized for the object classification sub-task, and we actually do not need any supervision for the object localization sub-task.

\textbf{Evaluation Metrics.} Since our localization method provides multiple bounding box outputs, and for fair comparison, we first only use our highest confident bounding box and follow \cite{CAM16} to adopt Top-1 localization (\textit{Top-1 Loc}), Top-5 localization (\textit{Top-5 Loc}), and GT-known localization accuracy (\textit{GT-known Loc}) for evaluation. 
Then we revise these criteria to take the multiple output bounding boxes into consideration.
\textit{Top-1 Loc} is unchanged, and it is correct when both the Top-1 classification and Top-1 localization results are correct. \textit{Top-5 Loc} is correct when one of the Top-5 predicted classes and one of the Top-5 predicted bounding boxes are both correct. \textit{GT-known Loc} is correct when there exists one predicted bounding box that has an IoU overlap over $0.5$ with any ground-truth bounding boxes, supposing that the ground-truth class is given. These results are reported separately and differentiated by underlining in the tables.

\begin{table*}[!t]
\small
    \centering
    \caption{Ablation study for the proposed binary-class detector (BCD) head and weighted entropy (WE) loss (\%). The results listed come from the Top-1 output bounding box, whereas the underlined results are from multiple output bounding boxes. $^{\star}$ denotes the re-implementation results of C$^{2}$AM \cite{C2AM22} by us.} 
    \setlength{\tabcolsep}{2.8mm}{
    \begin{tabular}{cccccccc} \toprule
    \multirow{2}{*}{Method} & \multirow{2}{*}{Loc Bac.} & \multicolumn{3}{c}{CUB-200-2011}     & \multicolumn{3}{c}{ImageNet-1K}      \\ \cmidrule(r){3-5} \cmidrule(r){6-8} 
                            &             & Top-1 Loc & Top-5 Loc & GT-known Loc & Top-1 Loc & Top-5 Loc & GT-known Loc \\ \midrule
    C$^{2}$AM \cite{C2AM22} & DenseNet161 & 81.76     & 91.11     & 92.88        & 59.56     & 67.05     & 68.53        \\ \midrule
    w/ Faster R-CNN         & DenseNet161 & 81.95     & 91.49 / \underline{91.74} & 93.36 / \underline{93.61} & 60.39     & 68.00 / \underline{69.92} & 69.51 / \underline{71.51} \\ 
    w/ Faster R-CNN+WE      & DenseNet161 & \textbf{83.24}     & \textbf{92.95} / \textbf{\underline{93.07}} & \textbf{94.87} / \textbf{\underline{94.99}} & \textbf{60.69}     & \textbf{68.36} / \textbf{\underline{70.80}} & \textbf{69.89} / \textbf{\underline{72.39}} \\ \midrule
    C$^{2}$AM \cite{C2AM22} & ResNet50    & 81.28$^{\star}$& 90.69$^{\star}$& 92.48$^{\star}$& 59.24$^{\star}$& 66.65$^{\star}$& 68.10$^{\star}$\\ \midrule
    w/ FCOS                 & ResNet50    & 82.70     & 92.33 / \underline{92.52} & 94.24 / \underline{94.42} & 60.11     & 67.69 / \underline{68.66} & 69.19 / \underline{70.18} \\
    w/ FCOS+WE              & ResNet50    & 83.05     & 92.71 / \underline{92.88} & 94.60 / \underline{94.77} & 60.18     & 67.76 / \underline{68.69} & 69.27 / \underline{70.22} \\
    w/ Faster R-CNN         & ResNet50    & 82.23     & 91.79 / \underline{91.92} & 93.63 / \underline{93.77} & 60.17     & 67.75 / \underline{70.08} & 69.24 / \underline{71.65} \\ 
    w/ Faster R-CNN+WE      & ResNet50    & 83.77     & 93.53 / \underline{93.84} & 95.47 / \underline{95.78} & 60.56     & 68.18 / \underline{70.66} & 69.67 / \underline{72.25} \\ 
    w/ CLN                  & ResNet50    & 82.87     & 92.43 / \underline{92.58} & 94.23 / \underline{94.41} & 60.68     & 68.30 / \underline{69.58} & 69.80 / \underline{71.14} \\
    w/ CLN+WE               & ResNet50    & \textbf{84.15}     & \textbf{93.98} / \underline{94.25} & \textbf{95.91} / \underline{96.19} & \textbf{60.70}     & \textbf{68.36} / \underline{69.65} & \textbf{69.86} / \underline{71.19} \\
    w/ Deformable DETR      & ResNet50    & 81.48     & 90.94 / \underline{95.54} & 92.81 / \underline{98.39} & 60.50     & 68.07 / \underline{78.64} & 69.58 / \underline{83.85} \\ 
    w/ Deformable DETR+WE   & ResNet50    & 82.32     & 91.83 / \textbf{\underline{95.86}} & 93.69 / \textbf{\underline{98.65}} & 60.57     & 68.16 / \textbf{\underline{79.04}} & 69.65 / \textbf{\underline{84.15}} \\ \bottomrule
    \end{tabular}}
    \label{tab:ab_2}
\end{table*}

The implementation details are deferred to the appendix.

\subsection{Results and Analyses}

In this section, we present comprehensive experimental results and perform analyses on these results to demonstrate the superiority of our method.

\begin{figure}[!t]
  \vspace{-0.1in}
  \centering
  \includegraphics[width=2.8in]{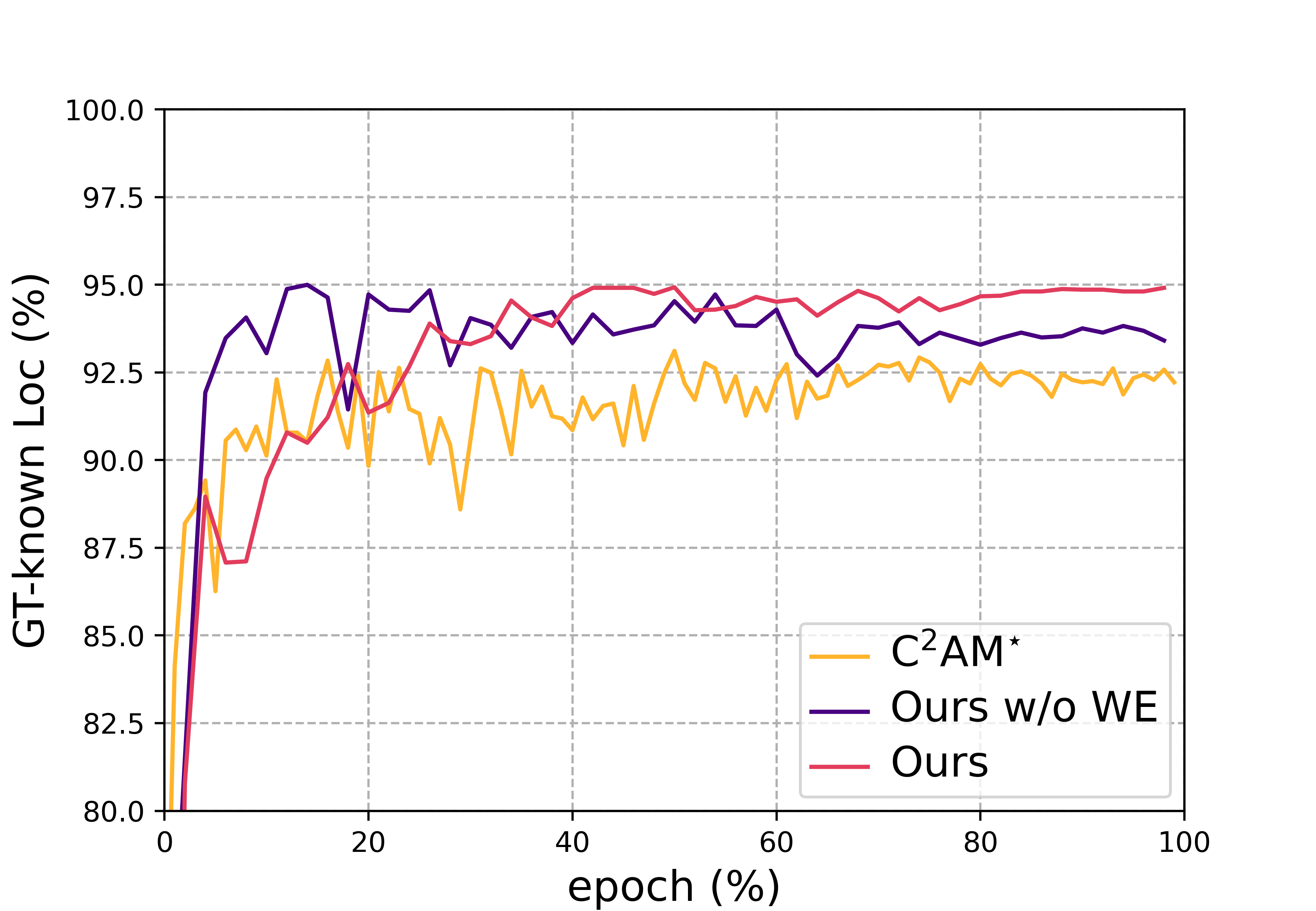}
  \caption{Top-1 localization result's \textit{GT-known Loc} accuracies on the CUB-200-2011 after each training epoch for the SOTA baseline and our methods (DenseNet161-Faster R-CNN). $^{\star}$ denotes the accuracy curve from our re-implementation.}
  \label{fig:ab_2_cub}
  \vspace{-0.1in}
\end{figure}

\begin{table}[!t]
\small
    \centering
    \caption{Ablation study for our WE loss using different $\gamma$ and $\alpha$ (\%) (DenseNet161-Faster R-CNN). $\eta$ is set to be $1$.}
    \setlength{\tabcolsep}{3.5mm}{
    \begin{tabular}{ccccc} \toprule
    \multirow{2}{*}{$\gamma$} & \multirow{2}{*}{$\alpha$} & \multicolumn{3}{c}{CUB-200-2011}  \\ \cmidrule(r){3-5} 
                              &                           & Top-1 Loc & Top-5 Loc & GT-known Loc \\ \midrule 
    2         & 0.5                & 81.73     & 91.12 / \underline{91.24} & 92.94 / \underline{93.06} \\ 
    2         & 0.25               & 82.13     & 91.63 / \underline{91.74} & 93.49 / \underline{93.61} \\ 
    4         & 0.25               & 82.23     & 91.74 / \underline{91.89} & 93.56 / \underline{93.71} \\ 
    4         & 0.1                & 82.25     & 91.77 / \underline{91.92} & 93.58 / \underline{93.77} \\ 
    6         & 0.1                & \textbf{82.31}     & \textbf{91.93} / \textbf{\underline{92.03}} & \textbf{93.79} / \textbf{\underline{93.90}} \\ 
    6         & 0.05               & 82.00     & 91.54 / \underline{91.66} & 93.41 / \underline{93.53} \\ \bottomrule
    \end{tabular}}
    \label{tab:ab_loss}
\end{table}

\begin{table}[!t]
\small
    \centering
    \caption{Ablation study for the proposed WEND using different $\eta$ (\%) (DenseNet161-Faster R-CNN). $\gamma$ and $\alpha$ are set to be 6 and 0.1 respectively.}
    \setlength{\tabcolsep}{4.2mm}{
    \begin{tabular}{cccc} \toprule
    \multirow{2}{*}{$\eta$}& \multicolumn{3}{c}{CUB-200-2011}  \\ \cmidrule(r){2-4} 
                           & Top-1 Loc & Top-5 Loc & GT-known Loc  \\ \midrule 
    4           & 82.20     & 91.81 / \underline{91.92} & 93.64 / \underline{93.76} \\ 
    2           & 82.08     & 91.61 / \underline{91.85} & 93.43 / \underline{93.67} \\ 
    1           & 82.31     & 91.93 / \underline{92.03} & 93.79 / \underline{93.90} \\ 
    0.5         & 82.26     & 91.83 / \underline{91.93} & 93.76 / \underline{93.85} \\ 
    0.25        & 82.90     & 92.54 / \underline{92.72} & 94.43 / \underline{94.60} \\ 
    0.125       & \textbf{83.24}     & \textbf{92.95} / \textbf{\underline{93.07}} & \textbf{94.87} / \textbf{\underline{94.99}} \\ 
    0.0625      & 82.61     & 92.22 / \underline{92.67} & 94.02 / \underline{94.49} \\ \bottomrule
    \end{tabular}}
    \label{tab:ab_b_cub}
\end{table}

\subsubsection{Ablation Study}
We first analyze the effectiveness of our proposed binary-class detector (BCD) and weighted entropy (WE) loss.
C$^2$AM \cite{C2AM22} is utilized as our baseline since we use the pseudo bounding boxes provided by it and the same backbones for classification and localization.
The results are reported in Table~\ref{tab:ab_2}. From the results, we observe that:
1) When using the DenseNet161 \cite{densenet17} as the localization backbone and the Faster R-CNN \cite{fasterrcnn15} based detector, the results on the CUB-200-2011 \cite{WahCUB_200_2011} are improved by over $1.3\%$ after applying our proposed WE loss. Meanwhile, the results on the ImageNet-1K \cite{ILSVRC15} are improved by around $1\%$ when evaluated using the Top-1 localization results, and up to about $3\%$ when evaluated using multiple localization results after applying the proposed detection head. The WE loss also improves the performance up to around $1\%$ when evaluated using multiple outputs. These results demonstrate that our proposed BCD and WE loss are beneficial to the WSOL task.
We also evaluate the Top-1 output bounding box in terms of the \textit{GT-known Loc} accuracy after each training epoch for dataset CUB-200-2011 \cite{WahCUB_200_2011}. From the results shown in Figure~\ref{fig:ab_2_cub}, we can see that BCD brings consistent performance improvements, while the WE loss leads to small perturbations at first and high performance in the end;
2) It is noted that the WE loss contributes more on the dataset CUB-200-2011 \cite{WahCUB_200_2011}, while the BCD is more critical on ImageNet-1K \cite{ILSVRC15}. This is because that CUB-200-2011 \cite{WahCUB_200_2011} is a fine-grained classification dataset and its low inter-class variance makes the discrimination between the foreground and the background more feasible and valuable. On the other hand, the object dataset ImageNet-1K \cite{ILSVRC15} vary in a wider range of sizes and numbers, making the flexible localization more significant;
3) We observe that although the Deformable DETR \cite{DeformableDETR21} usually yields more competitive performance than the Faster R-CNN \cite{fasterrcnn15}, it is less effective in our setting in terms of the Top-1 outputs due to the negative effect of the noisy pseudo bounding boxes. Yet its performance of the multiple outputs is quite excellent. This reveals that despite the inaccurate coordinates of the pseudo bounding boxes, the Deformable DETR \cite{DeformableDETR21} can still learn the object patterns. The label noise mainly confuses the classifier because the pseudo bounding boxes contain much background. Our proposed WE loss is able to deal with this problem to some extent.

\begin{figure}[!t]
  \centering
  \includegraphics[width=2.8in]{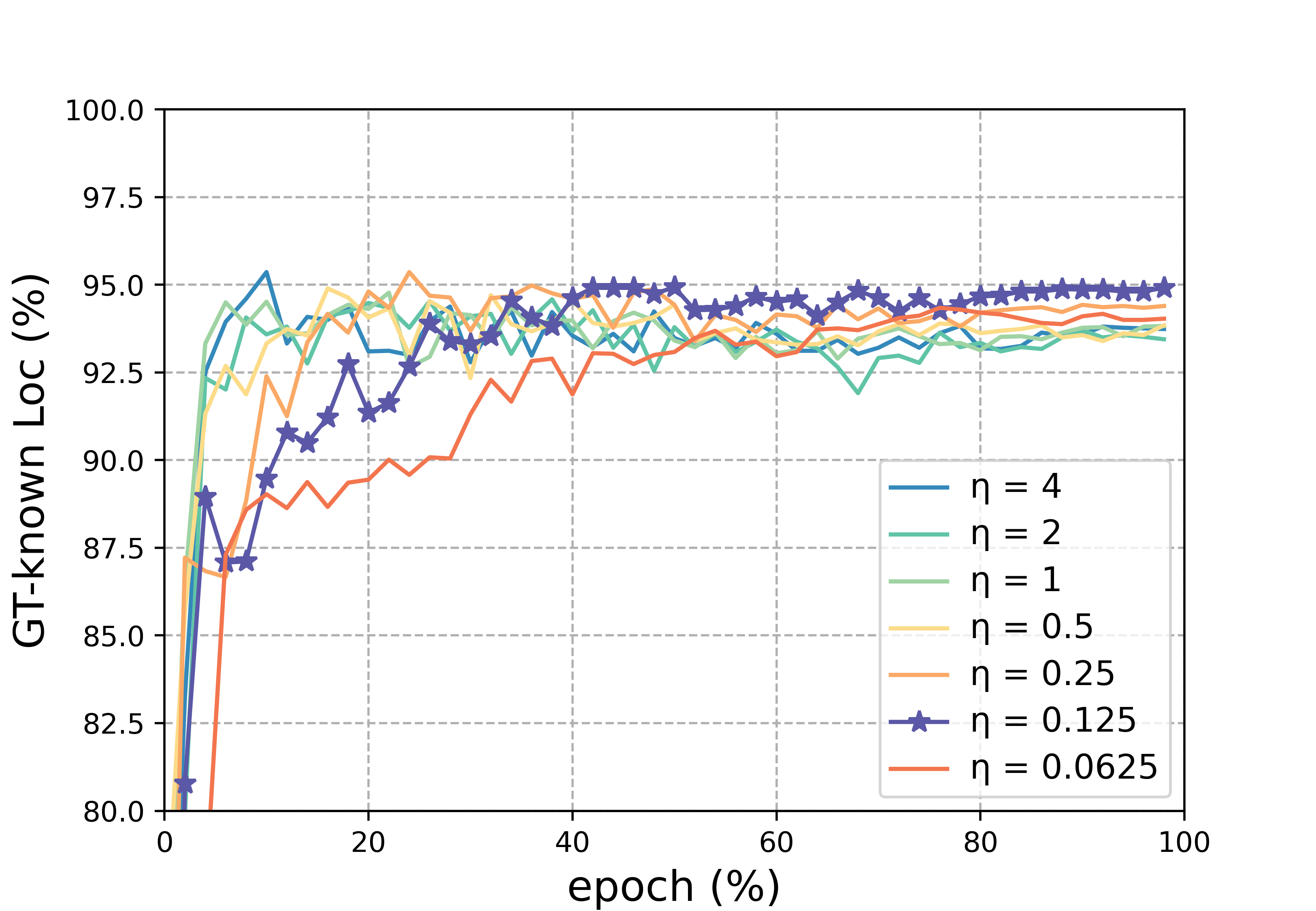}
  \caption{Top-1 localization result's \textit{GT-known Loc} accuracies on dataset CUB-200-2011 after each training epoch for the proposed WEND using different $\eta$ (DenseNet161-Faster R-CNN). $\gamma$ and $\alpha$ are set to be 6 and 0.1 respectively.}
  \label{fig:ab_b_cub}
  \vspace{-0.1in}
\end{figure}

\begin{table*}[!t]
\small
    \centering
    \caption{Comparison of our WEND and the SOTA approaches on the CUB-200-2011 test set and ImageNet-1K validation set (\%). Loc Bac. indicates the localization backbone, and Cls Bac. signifies the classification backbone. The results presented are from the Top-1 output bounding box, whereas the underlined results come from multiple output bounding boxes. $^{*}$ denotes the results provided in C$^{2}$AM \cite{C2AM22}. $^{\star}$ denotes the re-implementation results of C$^{2}$AM \cite{C2AM22} by us.}
    \setlength{\tabcolsep}{2.3mm}{
    \begin{tabular}{ccccccccc} \toprule
    \multirow{2}{*}{Method} & \multirow{2}{*}{Loc Bac.} & \multirow{2}{*}{Cls Bac.} & \multicolumn{3}{c}{CUB-200-2011}     & \multicolumn{3}{c}{ImageNet-1K}      \\ \cmidrule(r){4-6} \cmidrule(r){7-9} 
                            &                           &                           & Top-1 Loc & Top-5 Loc & GT-known Loc & Top-1 Loc & Top-5 Loc & GT-known Loc \\ \midrule
    CAM \cite{CAM16}        & \multicolumn{2}{c}{VGG16}                             & 44.15     & 52.16     & 56.0         & 42.80     & 54.86     & 59.00        \\ 
    CAM \cite{PSOL20}       & \multicolumn{2}{c}{DenseNet161}                       & 29.81     & 39.85     & -            & 39.61     & 50.40     & 52.54        \\
    ADL \cite{ADL19}        & \multicolumn{2}{c}{VGG16}                             & 52.36     & -         & -            & 44.92     & -         & -            \\ 
    I$^{2}$C \cite{I2C20}   & \multicolumn{2}{c}{InceptionV3}                       & 55.99     & 68.34     & 72.60        & 53.11     & 64.13     & 68.50        \\ 
    SPA \cite{SPA21}        & \multicolumn{2}{c}{VGG16}                             & 60.27     & 72.5      & 77.29        & 49.56     & 61.32     & 65.05        \\ 
    FAM \cite{FAM21}        & \multicolumn{2}{c}{VGG16}                             & 69.26     & -         & 89.26        & 51.96     & -         & 71.73        \\ 
    ORNet \cite{ORNet21}    & \multicolumn{2}{c}{VGG16}                             & 67.74     & 80.77     & 86.2         & 52.05     & 63.94     & 68.27        \\ 
    BAS \cite{cvpr22/BAS}   & \multicolumn{2}{c}{ResNet50}                          & 77.25     & 90.08     & 95.13         & 57.18     & \textbf{68.44}& \textbf{71.77}        \\ 
    PPD \cite{mm22/PPD}     & \multicolumn{2}{c}{ViT-S}                             & 78.8      & -         & \textbf{97.0}& 55.0      & -         & 67.5         \\ \midrule
    GC-Net \cite{GC-Net20}  & VGG16                     & VGG16                     & 63.24     & 75.54     & 81.10        & -         & -         & -            \\ 
    GC-Net \cite{GC-Net20}  & InceptionV3               & InceptionV3               & -         & -         & -            & 49.06     & 58.09     & -            \\ 
    PSOL \cite{PSOL20}      & ResNet50                  & ResNet50                  & 70.68     & 86.64     & 90.00$^{*}$  & 53.98     & 63.08     & 65.44        \\ 
    PSOL \cite{PSOL20}      & DenseNet161               & EfficientNet-B7           & 80.89$^{*}$& 89.97$^{*}$& 91.78$^{*}$& 58.00     & 65.02     & 66.28        \\ 
    SLT-Net \cite{SLTNet21} & ResNet50                  & ResNet50                  & 72.3      & -         & 90.7         & 56.2      & -         & 68.5         \\
    SLT-Net \cite{SLTNet21} & DenseNet161               & DenseNet161               & 75.8      & -         & 93.4         & 57.1      & -         & 69.0         \\
    C$^{2}$AM \cite{C2AM22} & ResNet50                  & EfficientNet-B7           & 81.28$^{\star}$& 90.69$^{\star}$& 92.48$^{\star}$& 59.24$^{\star}$& 66.65$^{\star}$& 68.10$^{\star}$\\
    C$^{2}$AM \cite{C2AM22} & DenseNet161               & EfficientNet-B7           & 81.76     & 91.11     & 92.88        & 59.56     & 67.05     & 68.53        \\ \midrule 
    WEND (Ours)             & ResNet50                  & EfficientNet-B7           & \textbf{83.77}     & \textbf{93.53} / \textbf{\underline{93.84}} & 95.47 / \underline{95.78} & 60.56     & 68.18 / \underline{70.66} & 69.67 / \underline{72.25} \\ 
    WEND (Ours)             & DenseNet161               & EfficientNet-B7           & 83.24     & 92.95 / \underline{93.07} & 94.87 / \underline{94.99} & \textbf{60.69}     & 68.36 / \underline{70.80} & 69.89 / \underline{72.39} \\ \bottomrule
    \end{tabular}}
    \label{tab:main_results}
\end{table*}


\begin{table*}[!t]
\small
    \centering
    \caption{Results of WEND's localization sub-part whose pre-trained backbone is freezed on the CUB-200-2011 test set and ImageNet-1K validation set (\%) (ResNet50). $^{\dag}$ denotes the localization sub-part initialized by the unsupervised pre-training (moco).}
    \setlength{\tabcolsep}{4.8mm}{
    \begin{tabular}{ccccccc} \toprule
    \multirow{2}{*}{Method} & \multicolumn{3}{c}{CUB-200-2011}     & \multicolumn{3}{c}{ImageNet-1K}      \\ \cmidrule(r){2-4} \cmidrule(r){5-7} 
                            & Top-1 Loc & Top-5 Loc & GT-known Loc & Top-1 Loc & Top-5 Loc & GT-known Loc \\ \midrule
    SCR                     & 70.37     & 77.80     & 79.17        & 56.14     & 63.34     & 64.77        \\
    SCR                     & 64.69$^{\dag}$& 71.89$^{\dag}$& 73.22$^{\dag}$ & 55.68$^{\dag}$& 62.80$^{\dag}$& 64.20$^{\dag}$ \\ \midrule
    FCOS                    & 79.01     & 88.04 / \underline{88.25} & 89.76 / \underline{89.97} & 57.78     & 65.15 / \underline{67.15} & 66.63 / \underline{68.72} \\ 
    FCOS                    & 58.75$^{\dag}$& 66.16 / \underline{69.78}$^{\dag}$& 67.65 / \underline{71.38}$^{\dag}$ & 58.75$^{\dag}$& 66.16 / \underline{69.78}$^{\dag}$& 67.65 / \underline{71.38}$^{\dag}$ \\ 
    Faster R-CNN            & 79.69     & 88.76 / \underline{89.72} & 90.39 / \underline{91.39} & 57.81     & 65.14 / \underline{68.29} & 66.61 / \underline{69.84} \\
    Faster R-CNN            & 68.30$^{\dag}$& 75.87 / \underline{77.66}$^{\dag}$& 77.42 / \underline{79.23}$^{\dag}$ & 58.33$^{\dag}$& 65.64 / \underline{65.92}$^{\dag}$& 67.12 / \underline{67.40}$^{\dag}$ \\
    CLN                     & 73.91     & 82.24 / \underline{82.22} & 83.79 / \underline{83.77} & 60.85     & 68.44 / \underline{69.16} & 69.94 / \underline{70.68}  \\ 
    CLN                     & 83.26$^{\dag}$& 92.85 / \underline{93.61}$^{\dag}$& 94.72 / \underline{95.47}$^{\dag}$ & 61.09$^{\dag}$& 68.72 / \underline{69.55}$^{\dag}$& 70.26 / \underline{71.11}$^{\dag}$ \\ 
    Deformable DETR         & 81.25     & 90.71 / \underline{96.45} & 92.56 / \underline{99.34} & 60.56     & 68.20 / \underline{79.27} & 69.71 / \underline{84.61} \\ 
    Deformable DETR         & 81.27$^{\dag}$& 90.82 / \underline{96.48}$^{\dag}$& 92.59 / \underline{99.19}$^{\dag}$& 60.59$^{\dag}$& 68.21 / \underline{79.29}$^{\dag}$& 69.72 / \underline{84.51}$^{\dag}$\\ \bottomrule
    \end{tabular}}
    \label{tab:freeze}
\end{table*}

\begin{figure*}[!t]
  \vspace{0.05in}
  \centering
  \includegraphics[width=6.8in]{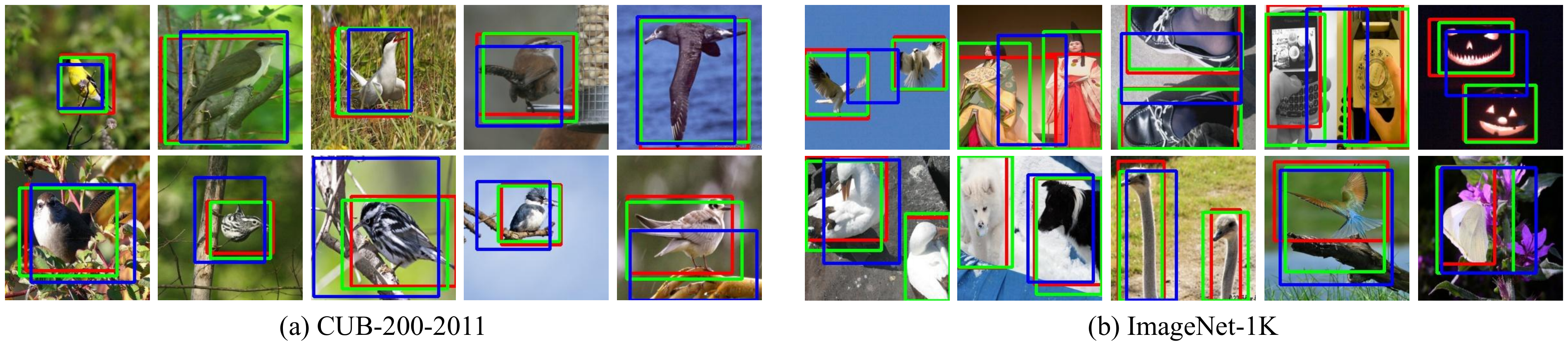}
  \caption{Visual comparison of our WEND and the C$^{2}$AM \cite{C2AM22} (DenseNet161). The red boxes, blue boxes, and green boxes denote the ground-truth bounding boxes, boxes predicted by the C$^{2}$AM \cite{C2AM22}, and boxes predicted by our method, respectively.}
  \label{fig:visual}
\end{figure*}


\subsubsection{Hyper-parameter Analysis}
We first analyze the choice of different hyper-parameters $\gamma$ and $\alpha$ in the WE loss.
Results on the CUB-200-2011 \cite{WahCUB_200_2011} for various combination of $\gamma$ and $\alpha$ are reported in Table~\ref{tab:ab_loss}. It can be seen from the results that the localization performance of our BCD is largely improved after applying the WE loss using proper hyper-parameters. Our method is insensitive to the change of $\gamma$ and $\alpha$.

Then we analyze the hyper-parameter $\eta$ for balancing of the pseudo supervised objective and the unsupervised objective.
Results for various $\eta$ are reported in Table~\ref{tab:ab_b_cub}. We can see that the performance first improve and then drop with an increasing $\eta$, and the best performance is achieved at $\eta=0.125$.

In addition, we check the Top-1 output bounding boxes' \textit{GT-known Loc} accuracies after each epoch of training for different $\eta$, and the results are plotted in Figure~\ref{fig:ab_b_cub}. It is observed that the \textit{GT-known Loc} curve of $\eta=0.125$ is smooth and works the best. The curves of $\eta$ larger than or equal to one are similar. Presumably, the big $\eta$ diminishes the role of the WE loss, and the results mostly come from the pseudo supervision. In contrast, smaller $\eta$ puts more weight on the unsupervised loss; the performance at the beginning is perturbed but it finally reaches a high level owing to the unsupervised loss' ability of handling with the noisy pseudo labels. This observation inspires us that we may combine the two losses in a better way to reduce the perturbation at the beginning of the training and further enhance the localization accuracy. We leave this as our future work.


\subsubsection{Comparison with Existing Approaches}

Table~\ref{tab:main_results} reports the comparison results of our proposed method and other approaches in WSOL on the two datasets.
The results are reported in Table~\ref{tab:main_results}, and it can be seen that our method consistently outperforms the existing works in terms of all criteria.

On the CUB-200-2011 \cite{WahCUB_200_2011}, our method 
outperforms the second best approach C$^{2}$AM \cite{C2AM22} by $2.01\%$, $2.42\%$, and $2.59\%$, respectively. When using all the output bounding boxes for evaluation, the performance of our method further improves, which outperforms the SLT-Net \cite{SLTNet21} by 2.38\% in terms of the \textit{GT-known Loc}. It is noteworthy that our method performs much better when using the ResNet50 \cite{resnet16} as the localization backbone than using DenseNet161 \cite{densenet17} on the CUB-200-2011 dataset. This is different from other separated localization-classification approaches. Since the last feature of the ResNet50 \cite{resnet16} has more clear semantic meaning than that of DenseNet161 \cite{densenet17}, this further indicates that our method can well reduce the negative effect of the noisy labels.

On the ImageNet-1K dataset, our method obtains $60.69\%$ \textit{Top-1 Loc}, $68.36\%$ \textit{Top-5 Loc}, and $69.89\%$ \textit{GT-known Loc} accuracies for the Top-1 class-agnostic bounding box output, which improves the C$^{2}$AM \cite{C2AM22} by $1.13\%$, $1.31\%$, and $1.36\%$, respectively. Since the images of ImageNet-1K \cite{ILSVRC15} may contain multiple ground-truth bounding boxes, our method performs much better when multiple output bounding boxes are considered for evaluation. In particular, our method achieves $70.80\%$ \textit{Top-5 Loc} and $72.39\%$ \textit{GT-known Loc}, which outperforms C$^{2}$AM \cite{C2AM22} by $3.75\%$ and $3.86\%$ respectively. These results demonstrate that our method significantly outperforms other approaches on different datasets.

\subsubsection{Feature Preservation.} We freeze the ImageNet pre-trained backbone weights to explore the impact of feature preservation for training using the noisy labels. Intuitively, the BCD of our proposed have higher model capacity than the simple SCR; and thus better performance should be achieved than SCR when the backbone is freezed.
Results reported in Table~\ref{tab:freeze} verify this intuition. Furthermore, comparing with the results of fine-tuning the backbone reported in Table~\ref{tab:ab_2}, our WEND (utilizing CLN and Deformable DETR) performs slightly better when the backbone is freezed on the ImageNet-1K dataset.
This reveals the potential of our method in enhancing the localization performance and reducing the computation cost to even less than that of the simple SCR (WEND-Deformable DETR + freeze backbone: 16.45M $\textless$ SCR: 24.55M).

\subsubsection{Visualization Comparison}

In Figure~\ref{fig:visual}, we visualize of the ground-truth bounding boxes, and the boxes predicted by the C$^{2}$AM \cite{C2AM22} counterpart and our WEND.
For the CUB-200-2011 dataset, it can be seen in Figure~\ref{fig:visual} (a) that the predicted boxes of our method are consistent with the ground truth. Yet the predicted boxes of the C$^{2}$AM \cite{C2AM22} either focuses on the discriminative parts or are affected by the confusing backgrounds. This indicates that our method has a better ability to distinguish the foreground objects and the backgrounds, thus predicting more complete and precise bounding box outputs. For the dataset ImageNet-1K \cite{ILSVRC15}, when it comes to some images with more than one ground truth boxes, C$^{2}$AM \cite{C2AM22} fails to predict the objects, while our method is able to correctly locates all the objects, as shown in the first row of Figure~\ref{fig:visual} (b). For some other images with multiple ground truths, the predicted boxes obtained by C$^{2}$AM have large offsets, while our results are more accurate. These results indicate that the performance of C$^{2}$AM \cite{C2AM22} is largely degraded when there are more than one object in the images. It may offset or completely mislocate the objects, since multiple objects confuse the SCR \cite{SCR14}. On the contrary, our method can provide multiple outputs to mitigate this confusion. Besides, our method also shows a better discrimination ability of the foreground and background on the ImageNet-1K dataset.

\section{Conclusions and Discussions}

In this paper, we rethink the localization of the state-of-the-art separated localization-classification pipeline in WSOL, and identify two major drawbacks: infeasibility of locating multiple objects and dealing with noisy pseudo labels.
To remedy these drawbacks, we propose a novel localization method termed WEND, which consists of a binary-class detector for predicting multiple outputs,
and an unsupervised weighted entropy (WE) loss to reduce the negative impact of the noisy labels. Comprehensive experiments show that our WEND significantly outperforms the existing approaches, and both the detection head and WE loss are beneficial and critical.

In the future, we intend to look deeper into the localization network design, and apply our method to more visual applications, e.g., video understanding, medical image analysis, and remote sensing imagery analysis.

%
\begin{acks}
This research was supported in part by the National Key Research and Development Program of China under No. 2021YFC3300200,
the Special Fund of Hubei Luojia Laboratory under Grant 220100014, the National Natural Science Foundation of China (Grant No. 62276195, 62141112, and 62225113),
the National Research Foundation Singapore and DSO National Laboratories under the AI Singapore Programme (AISG Award No: AISG2-GC-2023-006),
and the Fundamental Research Funds for the Central Universities (No. 2042023kf1033).
\end{acks}

\bibliographystyle{ACM-Reference-Format}
\bibliography{mm23_wend}


\begin{thebibliography}{55}


\ifx \showCODEN    \undefined \def \showCODEN     #1{\unskip}     \fi
\ifx \showDOI      \undefined \def \showDOI       #1{#1}\fi
\ifx \showISBNx    \undefined \def \showISBNx     #1{\unskip}     \fi
\ifx \showISBNxiii \undefined \def \showISBNxiii  #1{\unskip}     \fi
\ifx \showISSN     \undefined \def \showISSN      #1{\unskip}     \fi
\ifx \showLCCN     \undefined \def \showLCCN      #1{\unskip}     \fi
\ifx \shownote     \undefined \def \shownote      #1{#1}          \fi
\ifx \showarticletitle \undefined \def \showarticletitle #1{#1}   \fi
\ifx \showURL      \undefined \def \showURL       {\relax}        \fi
\providecommand\bibfield[2]{#2}
\providecommand\bibinfo[2]{#2}
\providecommand\natexlab[1]{#1}
\providecommand\showeprint[2][]{arXiv:#2}

\bibitem[Baek et~al\mbox{.}(2020)]%
        {PSY20}
\bibfield{author}{\bibinfo{person}{Kyungjune Baek}, \bibinfo{person}{Minhyun
  Lee}, {and} \bibinfo{person}{Hyunjung Shim}.}
  \bibinfo{year}{2020}\natexlab{}.
\newblock \showarticletitle{PsyNet: Self-Supervised Approach to Object
  Localization Using Point Symmetric Transformation}. In
  \bibinfo{booktitle}{\emph{AAAI}}. \bibinfo{pages}{10451--10459}.
\newblock


\bibitem[Berthelot et~al\mbox{.}(2019)]%
        {mixmatch19}
\bibfield{author}{\bibinfo{person}{David Berthelot}, \bibinfo{person}{Nicholas
  Carlini}, \bibinfo{person}{Ian~J. Goodfellow}, \bibinfo{person}{Nicolas
  Papernot}, \bibinfo{person}{Avital Oliver}, {and} \bibinfo{person}{Colin
  Raffel}.} \bibinfo{year}{2019}\natexlab{}.
\newblock \showarticletitle{MixMatch: {A} Holistic Approach to Semi-Supervised
  Learning}. In \bibinfo{booktitle}{\emph{NIPS}}. \bibinfo{pages}{5050--5060}.
\newblock


\bibitem[Carion et~al\mbox{.}(2020)]%
        {eccv20/detr}
\bibfield{author}{\bibinfo{person}{Nicolas Carion}, \bibinfo{person}{Francisco
  Massa}, \bibinfo{person}{Gabriel Synnaeve}, \bibinfo{person}{Nicolas
  Usunier}, \bibinfo{person}{Alexander Kirillov}, {and} \bibinfo{person}{Sergey
  Zagoruyko}.} \bibinfo{year}{2020}\natexlab{}.
\newblock \showarticletitle{End-to-End Object Detection with Transformers}. In
  \bibinfo{booktitle}{\emph{{ECCV}}}, Vol.~\bibinfo{volume}{12346}.
  \bibinfo{pages}{213--229}.
\newblock


\bibitem[Chen et~al\mbox{.}(2019)]%
        {ReDraw19}
\bibfield{author}{\bibinfo{person}{Micka{\"{e}}l Chen},
  \bibinfo{person}{Thierry Arti{\`{e}}res}, {and} \bibinfo{person}{Ludovic
  Denoyer}.} \bibinfo{year}{2019}\natexlab{}.
\newblock \showarticletitle{Unsupervised Object Segmentation by Redrawing}. In
  \bibinfo{booktitle}{\emph{NIPS}}. \bibinfo{pages}{12705--12716}.
\newblock


\bibitem[Cho et~al\mbox{.}(2015)]%
        {UODL15}
\bibfield{author}{\bibinfo{person}{Minsu Cho}, \bibinfo{person}{Suha Kwak},
  \bibinfo{person}{Cordelia Schmid}, {and} \bibinfo{person}{Jean Ponce}.}
  \bibinfo{year}{2015}\natexlab{}.
\newblock \showarticletitle{Unsupervised object discovery and localization in
  the wild: Part-based matching with bottom-up region proposals}. In
  \bibinfo{booktitle}{\emph{CVPR}}. \bibinfo{pages}{1201--1210}.
\newblock


\bibitem[Choe and Shim(2019)]%
        {ADL19}
\bibfield{author}{\bibinfo{person}{Junsuk Choe} {and} \bibinfo{person}{Hyunjung
  Shim}.} \bibinfo{year}{2019}\natexlab{}.
\newblock \showarticletitle{Attention-Based Dropout Layer for Weakly Supervised
  Object Localization}. In \bibinfo{booktitle}{\emph{CVPR}}.
  \bibinfo{pages}{2219--2228}.
\newblock


\bibitem[Dai et~al\mbox{.}(2021)]%
        {iccv21/dynamicDETR}
\bibfield{author}{\bibinfo{person}{Xiyang Dai}, \bibinfo{person}{Yinpeng Chen},
  \bibinfo{person}{Jianwei Yang}, \bibinfo{person}{Pengchuan Zhang},
  \bibinfo{person}{Lu Yuan}, {and} \bibinfo{person}{Lei Zhang}.}
  \bibinfo{year}{2021}\natexlab{}.
\newblock \showarticletitle{Dynamic {DETR:} End-to-End Object Detection with
  Dynamic Attention}. In \bibinfo{booktitle}{\emph{{ICCV}}}.
  \bibinfo{publisher}{{IEEE}}, \bibinfo{pages}{2968--2977}.
\newblock


\bibitem[Dhillon et~al\mbox{.}(2020)]%
        {fs20}
\bibfield{author}{\bibinfo{person}{Guneet~Singh Dhillon},
  \bibinfo{person}{Pratik Chaudhari}, \bibinfo{person}{Avinash Ravichandran},
  {and} \bibinfo{person}{Stefano Soatto}.} \bibinfo{year}{2020}\natexlab{}.
\newblock \showarticletitle{A Baseline for Few-Shot Image Classification}. In
  \bibinfo{booktitle}{\emph{ICLR}}.
\newblock


\bibitem[Grandvalet and Bengio(2004)]%
        {semi04}
\bibfield{author}{\bibinfo{person}{Yves Grandvalet} {and}
  \bibinfo{person}{Yoshua Bengio}.} \bibinfo{year}{2004}\natexlab{}.
\newblock \showarticletitle{Semi-supervised Learning by Entropy Minimization}.
  In \bibinfo{booktitle}{\emph{NIPS}}. \bibinfo{pages}{529--536}.
\newblock


\bibitem[Guo et~al\mbox{.}(2021)]%
        {SLTNet21}
\bibfield{author}{\bibinfo{person}{Guangyu Guo}, \bibinfo{person}{Junwei Han},
  \bibinfo{person}{Fang Wan}, {and} \bibinfo{person}{Dingwen Zhang}.}
  \bibinfo{year}{2021}\natexlab{}.
\newblock \showarticletitle{Strengthen Learning Tolerance for Weakly Supervised
  Object Localization}. In \bibinfo{booktitle}{\emph{CVPR}}.
  \bibinfo{pages}{7403--7412}.
\newblock


\bibitem[He et~al\mbox{.}(2020)]%
        {moco20}
\bibfield{author}{\bibinfo{person}{Kaiming He}, \bibinfo{person}{Haoqi Fan},
  \bibinfo{person}{Yuxin Wu}, \bibinfo{person}{Saining Xie}, {and}
  \bibinfo{person}{Ross~B. Girshick}.} \bibinfo{year}{2020}\natexlab{}.
\newblock \showarticletitle{Momentum Contrast for Unsupervised Visual
  Representation Learning}. In \bibinfo{booktitle}{\emph{CVPR}}.
  \bibinfo{pages}{9726--9735}.
\newblock


\bibitem[He et~al\mbox{.}(2016)]%
        {resnet16}
\bibfield{author}{\bibinfo{person}{Kaiming He}, \bibinfo{person}{Xiangyu
  Zhang}, \bibinfo{person}{Shaoqing Ren}, {and} \bibinfo{person}{Jian Sun}.}
  \bibinfo{year}{2016}\natexlab{}.
\newblock \showarticletitle{Deep Residual Learning for Image Recognition}. In
  \bibinfo{booktitle}{\emph{CVPR}}. \bibinfo{pages}{770--778}.
\newblock


\bibitem[Huang et~al\mbox{.}(2017)]%
        {densenet17}
\bibfield{author}{\bibinfo{person}{Gao Huang}, \bibinfo{person}{Zhuang Liu},
  \bibinfo{person}{Laurens van~der Maaten}, {and} \bibinfo{person}{Kilian~Q.
  Weinberger}.} \bibinfo{year}{2017}\natexlab{}.
\newblock \showarticletitle{Densely Connected Convolutional Networks}. In
  \bibinfo{booktitle}{\emph{CVPR}}. \bibinfo{pages}{2261--2269}.
\newblock


\bibitem[Kong et~al\mbox{.}(2020)]%
        {tip20/foveabox}
\bibfield{author}{\bibinfo{person}{Tao Kong}, \bibinfo{person}{Fuchun Sun},
  \bibinfo{person}{Huaping Liu}, \bibinfo{person}{Yuning Jiang},
  \bibinfo{person}{Lei Li}, {and} \bibinfo{person}{Jianbo Shi}.}
  \bibinfo{year}{2020}\natexlab{}.
\newblock \showarticletitle{FoveaBox: Beyound Anchor-Based Object Detection}.
\newblock \bibinfo{journal}{\emph{{IEEE} Trans. Image Process.}}
  \bibinfo{volume}{29} (\bibinfo{year}{2020}), \bibinfo{pages}{7389--7398}.
\newblock


\bibitem[Law and Deng(2018)]%
        {eccv18/cornernet}
\bibfield{author}{\bibinfo{person}{Hei Law} {and} \bibinfo{person}{Jia Deng}.}
  \bibinfo{year}{2018}\natexlab{}.
\newblock \showarticletitle{CornerNet: Detecting Objects as Paired Keypoints}.
  In \bibinfo{booktitle}{\emph{{ECCV}}}, Vol.~\bibinfo{volume}{11218}.
  \bibinfo{pages}{765--781}.
\newblock


\bibitem[Liang et~al\mbox{.}(2022)]%
        {shot22}
\bibfield{author}{\bibinfo{person}{Jian Liang}, \bibinfo{person}{Dapeng Hu},
  \bibinfo{person}{Yunbo Wang}, \bibinfo{person}{Ran He}, {and}
  \bibinfo{person}{Jiashi Feng}.} \bibinfo{year}{2022}\natexlab{}.
\newblock \showarticletitle{Source Data-Absent Unsupervised Domain Adaptation
  Through Hypothesis Transfer and Labeling Transfer}.
\newblock \bibinfo{journal}{\emph{{IEEE} Trans. Pattern Anal. Mach. Intell.}}
  \bibinfo{volume}{44}, \bibinfo{number}{11} (\bibinfo{year}{2022}),
  \bibinfo{pages}{8602--8617}.
\newblock


\bibitem[Lin et~al\mbox{.}(2017)]%
        {focalloss17}
\bibfield{author}{\bibinfo{person}{Tsung{-}Yi Lin}, \bibinfo{person}{Priya
  Goyal}, \bibinfo{person}{Ross~B. Girshick}, \bibinfo{person}{Kaiming He},
  {and} \bibinfo{person}{Piotr Doll{\'{a}}r}.} \bibinfo{year}{2017}\natexlab{}.
\newblock \showarticletitle{Focal Loss for Dense Object Detection}. In
  \bibinfo{booktitle}{\emph{ICCV}}. \bibinfo{pages}{2999--3007}.
\newblock


\bibitem[Liu et~al\mbox{.}(2021)]%
        {UT21}
\bibfield{author}{\bibinfo{person}{Yen{-}Cheng Liu},
  \bibinfo{person}{Chih{-}Yao Ma}, \bibinfo{person}{Zijian He},
  \bibinfo{person}{Chia{-}Wen Kuo}, \bibinfo{person}{Kan Chen},
  \bibinfo{person}{Peizhao Zhang}, \bibinfo{person}{Bichen Wu},
  \bibinfo{person}{Zsolt Kira}, {and} \bibinfo{person}{Peter Vajda}.}
  \bibinfo{year}{2021}\natexlab{}.
\newblock \showarticletitle{Unbiased Teacher for Semi-Supervised Object
  Detection}. In \bibinfo{booktitle}{\emph{ICLR}}.
\newblock


\bibitem[Lu et~al\mbox{.}(2020)]%
        {GC-Net20}
\bibfield{author}{\bibinfo{person}{Weizeng Lu}, \bibinfo{person}{Xi Jia},
  \bibinfo{person}{Weicheng Xie}, \bibinfo{person}{Linlin Shen},
  \bibinfo{person}{Yicong Zhou}, {and} \bibinfo{person}{Jinming Duan}.}
  \bibinfo{year}{2020}\natexlab{}.
\newblock \showarticletitle{Geometry Constrained Weakly Supervised Object
  Localization}. In \bibinfo{booktitle}{\emph{ECCV}},
  Vol.~\bibinfo{volume}{12371}. \bibinfo{pages}{481--496}.
\newblock


\bibitem[Meng et~al\mbox{.}(2021)]%
        {FAM21}
\bibfield{author}{\bibinfo{person}{Meng Meng}, \bibinfo{person}{Tianzhu Zhang},
  \bibinfo{person}{Qi Tian}, \bibinfo{person}{Yongdong Zhang}, {and}
  \bibinfo{person}{Feng Wu}.} \bibinfo{year}{2021}\natexlab{}.
\newblock \showarticletitle{Foreground Activation Maps for Weakly Supervised
  Object Localization}. In \bibinfo{booktitle}{\emph{ICCV}}.
  \bibinfo{pages}{3365--3375}.
\newblock


\bibitem[Oksuz et~al\mbox{.}(2021)]%
        {od_imbalance21}
\bibfield{author}{\bibinfo{person}{Kemal Oksuz}, \bibinfo{person}{Baris~Can
  Cam}, \bibinfo{person}{Sinan Kalkan}, {and} \bibinfo{person}{Emre Akbas}.}
  \bibinfo{year}{2021}\natexlab{}.
\newblock \showarticletitle{Imbalance Problems in Object Detection: {A}
  Review}.
\newblock \bibinfo{journal}{\emph{{IEEE} Trans. Pattern Anal. Mach. Intell.}}
  \bibinfo{volume}{43}, \bibinfo{number}{10} (\bibinfo{year}{2021}),
  \bibinfo{pages}{3388--3415}.
\newblock


\bibitem[Pan et~al\mbox{.}(2021)]%
        {SPA21}
\bibfield{author}{\bibinfo{person}{Xingjia Pan}, \bibinfo{person}{Yingguo Gao},
  \bibinfo{person}{Zhiwen Lin}, \bibinfo{person}{Fan Tang},
  \bibinfo{person}{Weiming Dong}, \bibinfo{person}{Haolei Yuan},
  \bibinfo{person}{Feiyue Huang}, {and} \bibinfo{person}{Changsheng Xu}.}
  \bibinfo{year}{2021}\natexlab{}.
\newblock \showarticletitle{Unveiling the Potential of Structure Preserving for
  Weakly Supervised Object Localization}. In \bibinfo{booktitle}{\emph{CVPR}}.
  \bibinfo{pages}{11642--11651}.
\newblock


\bibitem[Ren et~al\mbox{.}(2015)]%
        {fasterrcnn15}
\bibfield{author}{\bibinfo{person}{Shaoqing Ren}, \bibinfo{person}{Kaiming He},
  \bibinfo{person}{Ross~B. Girshick}, {and} \bibinfo{person}{Jian Sun}.}
  \bibinfo{year}{2015}\natexlab{}.
\newblock \showarticletitle{Faster {R-CNN:} Towards Real-Time Object Detection
  with Region Proposal Networks}. In \bibinfo{booktitle}{\emph{NIPS}}.
  \bibinfo{pages}{91--99}.
\newblock


\bibitem[Roh et~al\mbox{.}(2022)]%
        {iclr22/sparseDETR}
\bibfield{author}{\bibinfo{person}{Byungseok Roh}, \bibinfo{person}{Jaewoong
  Shin}, \bibinfo{person}{Wuhyun Shin}, {and} \bibinfo{person}{Saehoon Kim}.}
  \bibinfo{year}{2022}\natexlab{}.
\newblock \showarticletitle{Sparse {DETR:} Efficient End-to-End Object
  Detection with Learnable Sparsity}. In \bibinfo{booktitle}{\emph{{ICLR}}}.
\newblock


\bibitem[Russakovsky et~al\mbox{.}(2015)]%
        {ILSVRC15}
\bibfield{author}{\bibinfo{person}{Olga Russakovsky}, \bibinfo{person}{Jia
  Deng}, \bibinfo{person}{Hao Su}, \bibinfo{person}{Jonathan Krause},
  \bibinfo{person}{Sanjeev Satheesh}, \bibinfo{person}{Sean Ma},
  \bibinfo{person}{Zhiheng Huang}, \bibinfo{person}{Andrej Karpathy},
  \bibinfo{person}{Aditya Khosla}, \bibinfo{person}{Michael Bernstein},
  \bibinfo{person}{Alexander~C. Berg}, {and} \bibinfo{person}{Li Fei-Fei}.}
  \bibinfo{year}{2015}\natexlab{}.
\newblock \showarticletitle{{ImageNet Large Scale Visual Recognition
  Challenge}}.
\newblock \bibinfo{journal}{\emph{IJCV}} \bibinfo{volume}{115},
  \bibinfo{number}{3} (\bibinfo{year}{2015}), \bibinfo{pages}{211--252}.
\newblock


\bibitem[Sermanet et~al\mbox{.}(2014)]%
        {SCR14}
\bibfield{author}{\bibinfo{person}{Pierre Sermanet}, \bibinfo{person}{David
  Eigen}, \bibinfo{person}{Xiang Zhang}, \bibinfo{person}{Micha{\"{e}}l
  Mathieu}, \bibinfo{person}{Rob Fergus}, {and} \bibinfo{person}{Yann LeCun}.}
  \bibinfo{year}{2014}\natexlab{}.
\newblock \showarticletitle{OverFeat: Integrated Recognition, Localization and
  Detection using Convolutional Networks}. In \bibinfo{booktitle}{\emph{ICLR}}.
\newblock


\bibitem[Shannon(1948)]%
        {entropy1948}
\bibfield{author}{\bibinfo{person}{Claude~E. Shannon}.}
  \bibinfo{year}{1948}\natexlab{}.
\newblock \showarticletitle{A mathematical theory of communication}.
\newblock \bibinfo{journal}{\emph{Bell Syst. Tech. J.}} \bibinfo{volume}{27},
  \bibinfo{number}{3} (\bibinfo{year}{1948}), \bibinfo{pages}{379--423}.
\newblock


\bibitem[Simonyan and Zisserman(2015)]%
        {vgg15}
\bibfield{author}{\bibinfo{person}{Karen Simonyan} {and}
  \bibinfo{person}{Andrew Zisserman}.} \bibinfo{year}{2015}\natexlab{}.
\newblock \showarticletitle{Very Deep Convolutional Networks for Large-Scale
  Image Recognition}. In \bibinfo{booktitle}{\emph{ICLR}}.
\newblock


\bibitem[Su et~al\mbox{.}(2022)]%
        {JGP22}
\bibfield{author}{\bibinfo{person}{Yukun Su}, \bibinfo{person}{Guosheng Lin},
  \bibinfo{person}{Yun Hao}, \bibinfo{person}{Yiwen Cao},
  \bibinfo{person}{Wenjun Wang}, {and} \bibinfo{person}{Qingyao Wu}.}
  \bibinfo{year}{2022}\natexlab{}.
\newblock \showarticletitle{Self-Supervised Object Localization with Joint
  Graph Partition}. In \bibinfo{booktitle}{\emph{AAAI}}.
  \bibinfo{pages}{2289--2297}.
\newblock


\bibitem[Sun et~al\mbox{.}(2021)]%
        {cvpr21/sparsercnn}
\bibfield{author}{\bibinfo{person}{Peize Sun}, \bibinfo{person}{Rufeng Zhang},
  \bibinfo{person}{Yi Jiang}, \bibinfo{person}{Tao Kong},
  \bibinfo{person}{Chenfeng Xu}, \bibinfo{person}{Wei Zhan},
  \bibinfo{person}{Masayoshi Tomizuka}, \bibinfo{person}{Lei Li},
  \bibinfo{person}{Zehuan Yuan}, \bibinfo{person}{Changhu Wang}, {and}
  \bibinfo{person}{Ping Luo}.} \bibinfo{year}{2021}\natexlab{}.
\newblock \showarticletitle{Sparse {R-CNN:} End-to-End Object Detection With
  Learnable Proposals}. In \bibinfo{booktitle}{\emph{{CVPR}}}.
  \bibinfo{pages}{14454--14463}.
\newblock


\bibitem[Szegedy et~al\mbox{.}(2015)]%
        {googlenet15}
\bibfield{author}{\bibinfo{person}{Christian Szegedy}, \bibinfo{person}{Wei
  Liu}, \bibinfo{person}{Yangqing Jia}, \bibinfo{person}{Pierre Sermanet},
  \bibinfo{person}{Scott~E. Reed}, \bibinfo{person}{Dragomir Anguelov},
  \bibinfo{person}{Dumitru Erhan}, \bibinfo{person}{Vincent Vanhoucke}, {and}
  \bibinfo{person}{Andrew Rabinovich}.} \bibinfo{year}{2015}\natexlab{}.
\newblock \showarticletitle{Going deeper with convolutions}. In
  \bibinfo{booktitle}{\emph{CVPR}}. \bibinfo{pages}{1--9}.
\newblock


\bibitem[Szegedy et~al\mbox{.}(2016)]%
        {inception16}
\bibfield{author}{\bibinfo{person}{Christian Szegedy}, \bibinfo{person}{Vincent
  Vanhoucke}, \bibinfo{person}{Sergey Ioffe}, \bibinfo{person}{Jonathon
  Shlens}, {and} \bibinfo{person}{Zbigniew Wojna}.}
  \bibinfo{year}{2016}\natexlab{}.
\newblock \showarticletitle{Rethinking the Inception Architecture for Computer
  Vision}. In \bibinfo{booktitle}{\emph{CVPR}}. \bibinfo{pages}{2818--2826}.
\newblock


\bibitem[Tan and Le(2019)]%
        {efficientnet19}
\bibfield{author}{\bibinfo{person}{Mingxing Tan} {and} \bibinfo{person}{Quoc~V.
  Le}.} \bibinfo{year}{2019}\natexlab{}.
\newblock \showarticletitle{EfficientNet: Rethinking Model Scaling for
  Convolutional Neural Networks}. In \bibinfo{booktitle}{\emph{ICML}},
  Vol.~\bibinfo{volume}{97}. \bibinfo{pages}{6105--6114}.
\newblock


\bibitem[Tian et~al\mbox{.}(2019)]%
        {FCOS19}
\bibfield{author}{\bibinfo{person}{Zhi Tian}, \bibinfo{person}{Chunhua Shen},
  \bibinfo{person}{Hao Chen}, {and} \bibinfo{person}{Tong He}.}
  \bibinfo{year}{2019}\natexlab{}.
\newblock \showarticletitle{{FCOS:} Fully Convolutional One-Stage Object
  Detection}. In \bibinfo{booktitle}{\emph{{ICCV}}}.
  \bibinfo{pages}{9626--9635}.
\newblock


\bibitem[Vaswani et~al\mbox{.}(2017)]%
        {transformer17}
\bibfield{author}{\bibinfo{person}{Ashish Vaswani}, \bibinfo{person}{Noam
  Shazeer}, \bibinfo{person}{Niki Parmar}, \bibinfo{person}{Jakob Uszkoreit},
  \bibinfo{person}{Llion Jones}, \bibinfo{person}{Aidan~N. Gomez},
  \bibinfo{person}{Lukasz Kaiser}, {and} \bibinfo{person}{Illia Polosukhin}.}
  \bibinfo{year}{2017}\natexlab{}.
\newblock \showarticletitle{Attention is All you Need}. In
  \bibinfo{booktitle}{\emph{NIPS}}. \bibinfo{pages}{5998--6008}.
\newblock


\bibitem[Wah et~al\mbox{.}(2011)]%
        {WahCUB_200_2011}
\bibfield{author}{\bibinfo{person}{C. Wah}, \bibinfo{person}{S. Branson},
  \bibinfo{person}{P. Welinder}, \bibinfo{person}{P. Perona}, {and}
  \bibinfo{person}{S. Belongie}.} \bibinfo{year}{2011}\natexlab{}.
\newblock \bibinfo{booktitle}{\emph{{The Caltech-UCSD Birds-200-2011
  Dataset}}}.
\newblock \bibinfo{type}{{T}echnical {R}eport} CNS-TR-2011-001.
  \bibinfo{institution}{California Institute of Technology}.
\newblock


\bibitem[Wang et~al\mbox{.}(2021a)]%
        {tent21}
\bibfield{author}{\bibinfo{person}{Dequan Wang}, \bibinfo{person}{Evan
  Shelhamer}, \bibinfo{person}{Shaoteng Liu}, \bibinfo{person}{Bruno~A.
  Olshausen}, {and} \bibinfo{person}{Trevor Darrell}.}
  \bibinfo{year}{2021}\natexlab{a}.
\newblock \showarticletitle{Tent: Fully Test-Time Adaptation by Entropy
  Minimization}. In \bibinfo{booktitle}{\emph{ICLR}}.
\newblock


\bibitem[Wang et~al\mbox{.}(2021b)]%
        {ijcai21/SAIL}
\bibfield{author}{\bibinfo{person}{Yunke Wang}, \bibinfo{person}{Chang Xu},
  {and} \bibinfo{person}{Bo Du}.} \bibinfo{year}{2021}\natexlab{b}.
\newblock \showarticletitle{Robust Adversarial Imitation Learning via
  Adaptively-Selected Demonstrations}. In \bibinfo{booktitle}{\emph{{IJCAI}}}.
  \bibinfo{pages}{3155--3161}.
\newblock


\bibitem[Wang et~al\mbox{.}(2023)]%
        {Unidetector23}
\bibfield{author}{\bibinfo{person}{Zhenyu Wang}, \bibinfo{person}{Yali Li},
  \bibinfo{person}{Xi Chen}, \bibinfo{person}{Ser{-}Nam Lim},
  \bibinfo{person}{Antonio Torralba}, \bibinfo{person}{Hengshuang Zhao}, {and}
  \bibinfo{person}{Shengjin Wang}.} \bibinfo{year}{2023}\natexlab{}.
\newblock \showarticletitle{Detecting Everything in the Open World: Towards
  Universal Object Detection}.
\newblock \bibinfo{journal}{\emph{CoRR}}  \bibinfo{volume}{abs/2303.11749}
  (\bibinfo{year}{2023}).
\newblock


\bibitem[Wei et~al\mbox{.}(2017)]%
        {SCDA17}
\bibfield{author}{\bibinfo{person}{Xiu{-}Shen Wei}, \bibinfo{person}{Jian{-}Hao
  Luo}, \bibinfo{person}{Jianxin Wu}, {and} \bibinfo{person}{Zhi{-}Hua Zhou}.}
  \bibinfo{year}{2017}\natexlab{}.
\newblock \showarticletitle{Selective Convolutional Descriptor Aggregation for
  Fine-Grained Image Retrieval}.
\newblock \bibinfo{journal}{\emph{{IEEE} Trans. Image Process.}}
  \bibinfo{volume}{26}, \bibinfo{number}{6} (\bibinfo{year}{2017}),
  \bibinfo{pages}{2868--2881}.
\newblock


\bibitem[Wei et~al\mbox{.}(2019)]%
        {DDT19}
\bibfield{author}{\bibinfo{person}{Xiu{-}Shen Wei}, \bibinfo{person}{Chen{-}Lin
  Zhang}, \bibinfo{person}{Jianxin Wu}, \bibinfo{person}{Chunhua Shen}, {and}
  \bibinfo{person}{Zhi{-}Hua Zhou}.} \bibinfo{year}{2019}\natexlab{}.
\newblock \showarticletitle{Unsupervised object discovery and co-localization
  by deep descriptor transformation}.
\newblock \bibinfo{journal}{\emph{Pattern Recognit.}}  \bibinfo{volume}{88}
  (\bibinfo{year}{2019}), \bibinfo{pages}{113--126}.
\newblock


\bibitem[Wu et~al\mbox{.}(2022)]%
        {cvpr22/BAS}
\bibfield{author}{\bibinfo{person}{Pingyu Wu}, \bibinfo{person}{Wei Zhai},
  {and} \bibinfo{person}{Yang Cao}.} \bibinfo{year}{2022}\natexlab{}.
\newblock \showarticletitle{Background Activation Suppression for Weakly
  Supervised Object Localization}. In \bibinfo{booktitle}{\emph{{CVPR}}}.
  \bibinfo{pages}{14228--14237}.
\newblock


\bibitem[Xie et~al\mbox{.}(2021)]%
        {ORNet21}
\bibfield{author}{\bibinfo{person}{Jinheng Xie}, \bibinfo{person}{Cheng Luo},
  \bibinfo{person}{Xiangping Zhu}, \bibinfo{person}{Ziqi Jin},
  \bibinfo{person}{Weizeng Lu}, {and} \bibinfo{person}{Linlin Shen}.}
  \bibinfo{year}{2021}\natexlab{}.
\newblock \showarticletitle{Online Refinement of Low-level Feature Based
  Activation Map for Weakly Supervised Object Localization}. In
  \bibinfo{booktitle}{\emph{ICCV}}. \bibinfo{pages}{132--141}.
\newblock


\bibitem[Xie et~al\mbox{.}(2022)]%
        {C2AM22}
\bibfield{author}{\bibinfo{person}{Jinheng Xie}, \bibinfo{person}{Jianfeng
  Xiang}, \bibinfo{person}{Junliang Chen}, \bibinfo{person}{Xianxu Hou},
  \bibinfo{person}{Xiaodong Zhao}, {and} \bibinfo{person}{Linlin Shen}.}
  \bibinfo{year}{2022}\natexlab{}.
\newblock \showarticletitle{C\({}^{\mbox{2}}\) {AM:} Contrastive learning of
  Class-agnostic Activation Map for Weakly Supervised Object Localization and
  Semantic Segmentation}. In \bibinfo{booktitle}{\emph{CVPR}}.
  \bibinfo{pages}{979--988}.
\newblock


\bibitem[Xu et~al\mbox{.}(2015)]%
        {ijcai15/SPL}
\bibfield{author}{\bibinfo{person}{Chang Xu}, \bibinfo{person}{Dacheng Tao},
  {and} \bibinfo{person}{Chao Xu}.} \bibinfo{year}{2015}\natexlab{}.
\newblock \showarticletitle{Multi-view Self-Paced Learning for Clustering}. In
  \bibinfo{booktitle}{\emph{{IJCAI}}}. \bibinfo{pages}{3974--3980}.
\newblock


\bibitem[Xu et~al\mbox{.}(2022b)]%
        {mm22/PPD}
\bibfield{author}{\bibinfo{person}{Jingyuan Xu}, \bibinfo{person}{Hongtao Xie},
  \bibinfo{person}{Chuanbin Liu}, {and} \bibinfo{person}{Yongdong Zhang}.}
  \bibinfo{year}{2022}\natexlab{b}.
\newblock \showarticletitle{Proxy Probing Decoder for Weakly Supervised Object
  Localization: {A} Baseline Investigation}. In \bibinfo{booktitle}{\emph{{ACM
  MM}}}. \bibinfo{pages}{4185--4193}.
\newblock


\bibitem[Xu et~al\mbox{.}(2022a)]%
        {miccai22/lssanet}
\bibfield{author}{\bibinfo{person}{Rui Xu}, \bibinfo{person}{Yong Luo},
  \bibinfo{person}{Bo Du}, \bibinfo{person}{Kaiming Kuang}, {and}
  \bibinfo{person}{Jiancheng Yang}.} \bibinfo{year}{2022}\natexlab{a}.
\newblock \showarticletitle{LSSANet: {A} Long Short Slice-Aware Network for
  Pulmonary Nodule Detection}. In \bibinfo{booktitle}{\emph{{MICCAI}}},
  Vol.~\bibinfo{volume}{13431}. \bibinfo{pages}{664--674}.
\newblock


\bibitem[Zhang et~al\mbox{.}(2020a)]%
        {PSOL20}
\bibfield{author}{\bibinfo{person}{Chen{-}Lin Zhang},
  \bibinfo{person}{Yun{-}Hao Cao}, {and} \bibinfo{person}{Jianxin Wu}.}
  \bibinfo{year}{2020}\natexlab{a}.
\newblock \showarticletitle{Rethinking the Route Towards Weakly Supervised
  Object Localization}. In \bibinfo{booktitle}{\emph{CVPR}}.
  \bibinfo{pages}{13457--13466}.
\newblock


\bibitem[Zhang et~al\mbox{.}(2018)]%
        {DUSD18}
\bibfield{author}{\bibinfo{person}{Jing Zhang}, \bibinfo{person}{Tong Zhang},
  \bibinfo{person}{Yuchao Dai}, \bibinfo{person}{Mehrtash Harandi}, {and}
  \bibinfo{person}{Richard~I. Hartley}.} \bibinfo{year}{2018}\natexlab{}.
\newblock \showarticletitle{Deep Unsupervised Saliency Detection: {A} Multiple
  Noisy Labeling Perspective}. In \bibinfo{booktitle}{\emph{CVPR}}.
  \bibinfo{pages}{9029--9038}.
\newblock


\bibitem[Zhang et~al\mbox{.}(2019)]%
        {MO19}
\bibfield{author}{\bibinfo{person}{Runsheng Zhang}, \bibinfo{person}{Yaping
  Huang}, \bibinfo{person}{Mengyang Pu}, \bibinfo{person}{Qingji Guan},
  \bibinfo{person}{Jian Zhang}, {and} \bibinfo{person}{Qi Zou}.}
  \bibinfo{year}{2019}\natexlab{}.
\newblock \showarticletitle{Mining Objects: Fully Unsupervised Object Discovery
  and Localization From a Single Image}.
\newblock \bibinfo{journal}{\emph{CoRR}}  \bibinfo{volume}{abs/1902.09968}
  (\bibinfo{year}{2019}).
\newblock


\bibitem[Zhang et~al\mbox{.}(2020b)]%
        {cvpr20/atss}
\bibfield{author}{\bibinfo{person}{Shifeng Zhang}, \bibinfo{person}{Cheng Chi},
  \bibinfo{person}{Yongqiang Yao}, \bibinfo{person}{Zhen Lei}, {and}
  \bibinfo{person}{Stan~Z. Li}.} \bibinfo{year}{2020}\natexlab{b}.
\newblock \showarticletitle{Bridging the Gap Between Anchor-Based and
  Anchor-Free Detection via Adaptive Training Sample Selection}. In
  \bibinfo{booktitle}{\emph{{CVPR}}}. \bibinfo{pages}{9756--9765}.
\newblock


\bibitem[Zhang et~al\mbox{.}(2020c)]%
        {I2C20}
\bibfield{author}{\bibinfo{person}{Xiaolin Zhang}, \bibinfo{person}{Yunchao
  Wei}, {and} \bibinfo{person}{Yi Yang}.} \bibinfo{year}{2020}\natexlab{c}.
\newblock \showarticletitle{Inter-Image Communication for Weakly Supervised
  Localization}. In \bibinfo{booktitle}{\emph{ECCV}},
  Vol.~\bibinfo{volume}{12364}. \bibinfo{pages}{271--287}.
\newblock


\bibitem[Zhao et~al\mbox{.}(2021)]%
        {DiLo21}
\bibfield{author}{\bibinfo{person}{Nanxuan Zhao}, \bibinfo{person}{Zhirong Wu},
  \bibinfo{person}{Rynson W.~H. Lau}, {and} \bibinfo{person}{Stephen Lin}.}
  \bibinfo{year}{2021}\natexlab{}.
\newblock \showarticletitle{Distilling Localization for Self-Supervised
  Representation Learning}. In \bibinfo{booktitle}{\emph{AAAI}}.
  \bibinfo{pages}{10990--10998}.
\newblock


\bibitem[Zhou et~al\mbox{.}(2016)]%
        {CAM16}
\bibfield{author}{\bibinfo{person}{Bolei Zhou}, \bibinfo{person}{Aditya
  Khosla}, \bibinfo{person}{{\`{A}}gata Lapedriza}, \bibinfo{person}{Aude
  Oliva}, {and} \bibinfo{person}{Antonio Torralba}.}
  \bibinfo{year}{2016}\natexlab{}.
\newblock \showarticletitle{Learning Deep Features for Discriminative
  Localization}. In \bibinfo{booktitle}{\emph{CVPR}}.
  \bibinfo{pages}{2921--2929}.
\newblock


\bibitem[Zhu et~al\mbox{.}(2021)]%
        {DeformableDETR21}
\bibfield{author}{\bibinfo{person}{Xizhou Zhu}, \bibinfo{person}{Weijie Su},
  \bibinfo{person}{Lewei Lu}, \bibinfo{person}{Bin Li},
  \bibinfo{person}{Xiaogang Wang}, {and} \bibinfo{person}{Jifeng Dai}.}
  \bibinfo{year}{2021}\natexlab{}.
\newblock \showarticletitle{Deformable {DETR:} Deformable Transformers for
  End-to-End Object Detection}. In \bibinfo{booktitle}{\emph{ICLR}}.
\newblock


\end{thebibliography}

\clearpage

\appendix

\section{Implementation Details}

For our method, we directly use the pre-trained EfficientNet \cite{efficientnet19} for the object classification sub-task\footnote{The pre-trained weights for dataset CUB-200-2011 are downloaded from \url{https://github.com/CVI-SZU/CCAM}}. In regard to the object localization sub-task, we use the generated pseudo bounding boxes provided in C$^{2}$AM \cite{C2AM22} for training the detector. We choose the ResNet50 \cite{resnet16} and DenseNet161 \cite{densenet17} pre-trained on the ImageNet as our backbone, and implement the object detectors of various types, the popular single-stage FCOS \cite{FCOS19}, two-stage Faster R-CNN \cite{fasterrcnn15} and CLN \cite{Unidetector23}, and recent transformer-based Deformable DETR \cite{DeformableDETR21}. Our proposed weighted entropy (WE) loss is implemented in the classifier of the FCOS \cite{FCOS19} and CLN \cite{Unidetector23}, the classifier of the R-CNN \cite{UT21} for the Faster R-CNN \cite{fasterrcnn15}, and incorporated in every stage of the iterative loss function for the Deformable DETR \cite{DeformableDETR21}. 
The input images are first resized to 256$\times$256. Then during training, we randomly crop the images to 224$\times$224 and apply horizontal flipping for data augmentation. For testing, the images are center cropped to 224$\times$224. All images are normalized.

For all the experiments, we train our localization model 50 epochs on the CUB-200-2011 \cite{WahCUB_200_2011} and 10 epochs on the ImageNet-1K \cite{ILSVRC15} with a training batch size of 256. For the FCOS \cite{FCOS19}, Faster R-CNN \cite{fasterrcnn15} and CLN \cite{Unidetector23}, the Stochastic Gradient Descent (SGD) optimizer is used. The momentum and the weight decay are set to 0.9 and 0.0001 respectively. The initialization learning rate is 0.004 for the backbone and doubled for the detection head; the cosine annealing policy is applied to schedule the learning rate. For the Deformable DETR \cite{DeformableDETR21}, the Adam optimizer is used with $\beta_{1}=0.9$, $\beta_{2}=0.999$, and weight decay of 0.0001, respectively. The initialization learning rate is $2\times10^{-4}$, and the learning rates of the backbone and the linear projections, which are used for predicting object query reference and sampling offsets, are reduced to 0.1; all learning rates are scheduled using the cosine annealing policy. Our implementation is in PyTorch on 4 NVIDIA GeForce RTX 3090 GPUs with 24GB memory. 

\section{Additional Experiments}


\subsection{Strict Unsupervised Localization}

Since we adopt the pseudo bounding boxes generated via unsupervised learning \cite{C2AM22}, our localization sub-part can be converted to a strict unsupervised setting by initializing the localization backbone using the unsupervised pre-training, moco \cite{moco20}. We compare our strict unsupervised localization sub-part with C$^2$AM \cite{C2AM22}, and the results are reported in~Table~\ref{tab:unsup_main_results}. It can be seen that our method also outperforms C$^2$AM \cite{C2AM22} in this strict unsupervised setting, and especially achieves a significant $3\%$ improvement on the CUB-200-2011 \cite{WahCUB_200_2011} dataset.


\begin{table}[!t]
    \small
    \centering
    \caption{The GT-known Loc of our WEND's localization sub-part initialized by the unsupervised pre-training (moco) and the unsupervised C$^2$AM \cite{C2AM22} on the CUB-200-2011 test set and ImageNet-1K validation set (\%). The results in the first line are performance of C$^2$AM \cite{C2AM22} utilizing moco to generate the pseudo bounding boxes; $^{\star}$ denotes the re-implementation results of the C$^{2}$AM \cite{C2AM22} by us, which uses moco to initialize the localization network.}
    \setlength{\tabcolsep}{3.3mm}{
    \begin{tabular}{cccc} \toprule
    Method                  & Loc Bac.                          & CUB-200-2011   & ImageNet-1K    \\ \midrule
    C$^{2}$AM \cite{C2AM22} & ResNet50                          & 89.90          & 66.51          \\ 
    C$^{2}$AM \cite{C2AM22} & ResNet50                          & 91.97$^{\star}$& 68.44$^{\star}$\\ \midrule
    Faster R-CNN            & ResNet50                          & \textbf{94.59} / \underline{94.97} & 69.65 / \underline{71.36}     \\ 
    CLN                     & ResNet50                          & 94.48 / \underline{94.95} & \textbf{69.92} / \underline{71.42}     \\ 
    Deformable DETR         & ResNet50                          & 93.00 / \textbf{\underline{97.84}} & 69.52 / \textbf{\underline{84.11}}     \\ 
    \bottomrule
    \end{tabular}}
    \label{tab:unsup_main_results}
\end{table}

\begin{table}[!t]
    \small
    \centering
    \caption{Ablation study for the proposed WEND using different $\eta$ (\%) (DenseNet161-Faster R-CNN); $\gamma$, $\alpha$, and $\tau$ are set to be $4$, $0.25$, and $0.1$ respectively.}
    \setlength{\tabcolsep}{4mm}{
    \begin{tabular}{cccc} \toprule
    \multirow{2}{*}{$\eta$}& \multicolumn{3}{c}{ImageNet-1K}  \\ \cmidrule(r){2-4} 
                           & Top-1 Loc & Top-5 Loc & GT-known Loc  \\ \midrule 
    4           & 60.49     & 68.11 / \underline{70.07}& 69.63 / \underline{71.66}\\ 
    1           & 60.51     & 68.12 / \underline{70.04}& 69.64 / \underline{71.63}\\ 
    0.5         & 60.43     & 68.02 / \underline{69.99}& 69.53 / \underline{71.57}\\ 
    0.125       & 60.39     & 68.01 / \underline{70.02}& 69.53 / \underline{71.60}\\ 
    0.0625      & 60.55     & 68.18 / \textbf{\underline{70.82}}& 69.70 / \textbf{\underline{72.42}}\\ 
    0.03125     & \textbf{60.69}     & \textbf{68.36} / \underline{70.80}& \textbf{69.89} / \underline{72.39}\\ 
    0.015625    & 60.46     & 68.10 / \underline{70.16}& 69.63 / \underline{71.76}\\ \bottomrule
    \end{tabular}}
    \label{tab:ab_b_imagenet}
\end{table}

\begin{table}[!t]
    \small
    \centering
    \caption{Ablation study for the proposed WEND using different $\tau$ (\%) (DenseNet161-Faster R-CNN), where $\gamma$, $\alpha$, and $\eta$ are set to be $6$, $0.1$, and $0.125$ respectively.}
    \setlength{\tabcolsep}{4.7mm}{
    \begin{tabular}{cccc} \toprule
    \multirow{2}{*}{$\tau$}& \multicolumn{3}{c}{CUB-200-2011}  \\ \cmidrule(r){2-4} 
                           & Top-1 Loc & Top-5 Loc & GT-known Loc  \\ \midrule 
    0.1         & 83.17     & 92.82 / \underline{93.05}& 94.72 / \underline{94.98}\\ 
    0.2         & 83.13     & 92.78 / \underline{92.96}& 94.66 / \underline{94.85}\\ 
    0.3         & 83.24     & \textbf{92.95} / \textbf{\underline{93.07}}& \textbf{94.87} / \textbf{\underline{94.99}}\\ 
    0.4         & 83.04     & 92.70 / \underline{92.87}& 94.59 / \underline{94.78}\\ 
    0.5         & \textbf{83.29}     & \textbf{92.95} / \underline{93.03}& 94.85 / \underline{94.94}\\  \bottomrule
    \end{tabular}}
    \label{tab:ab_thres_cub}
\end{table}

\begin{table*}[t]
    \small
    \centering
    \caption{Ablation study for the standard entropy loss and the proposed weighted entropy (WE) loss (\%) (DenseNet161-Faster R-CNN). The baseline is the binary-class detector (BCD), $\eta$ is set to be $0.125$ for CUB-200-2011 and $0.03125$ for ImageNet-1K.}
    \setlength{\tabcolsep}{5.5mm}{
    \begin{tabular}{ccccccc} \toprule
    \multirow{2}{*}{Method} & \multicolumn{3}{c}{CUB-200-2011}     & \multicolumn{3}{c}{ImageNet-1K}      \\ \cmidrule(r){2-4} \cmidrule(r){5-7} 
                            & Top-1 Loc & Top-5 Loc & GT-known Loc & Top-1 Loc & Top-5 Loc & GT-known Loc \\ \midrule 
    BCD                     & 81.95     & 91.49 / \underline{91.74}& 93.36 / \underline{93.61}& 60.39     & 68.00 / \underline{69.92}& 69.51 / \underline{71.51}\\ 
    w/ Entropy              & 82.83     & 92.42 / \underline{92.85}& 94.28 / \underline{94.72}& 60.65     & 68.30 / \underline{70.62}& 69.84 / \underline{72.23}\\ 
    w/ WE                   & \textbf{83.24}     & \textbf{92.95} / \textbf{\underline{93.07}}& \textbf{94.87} / \textbf{\underline{94.99}}& \textbf{60.69}     & \textbf{68.36} / \textbf{\underline{70.80}}& \textbf{69.89} / \textbf{\underline{72.39}}      \\ \bottomrule
    \end{tabular}}
    \label{tab:ab_entropy}
\end{table*}

\begin{table}[!t]
    \small
    \centering
    \caption{Ablation study for the proposed WEND using different $\tau$ (\%) (ResNet50-Deformable DETR), where $\gamma$, $\alpha$, and $\eta$ are set to be $6$, $0.1$, and $1$ respectively.}
    \setlength{\tabcolsep}{4.7mm}{
    \begin{tabular}{cccc} \toprule
    \multirow{2}{*}{$\tau$}& \multicolumn{3}{c}{CUB-200-2011}  \\ \cmidrule(r){2-4} 
                           & Top-1 Loc & Top-5 Loc & GT-known Loc  \\ \midrule 
    0.5         & 81.44     & 90.99 / \textbf{\underline{95.70}}& 92.81 / \textbf{\underline{98.78}}\\ 
    0.7         & 81.52     & 90.99 / \underline{95.52}& 92.86 / \underline{98.45}\\ 
    0.9         & \textbf{82.13}     & \textbf{91.64} / \underline{95.67}& \textbf{93.43} / \underline{98.47}\\  \bottomrule
    \end{tabular}}
    \label{tab:ab_thres_cub_detr}
\end{table}

\begin{table}[!t]
    \small
    \centering
    \caption{Ablation study for the proposed WEND using different positive-to-negative ratios (\%) (DenseNet161-Faster R-CNN), where $\gamma$, $\alpha$, $\eta$, and $\tau$ are set to be $6$, $0.1$, $0.125$, and $0.3$ respectively.}
    \setlength{\tabcolsep}{4.7mm}{
    \begin{tabular}{cccc} \toprule
    \multirow{2}{*}{$r$}& \multicolumn{3}{c}{CUB-200-2011}  \\ \cmidrule(r){2-4} 
                           & Top-1 Loc & Top-5 Loc & GT-known Loc  \\ \midrule 
    1:1         & 83.24     & 92.95 / \underline{93.07}& 94.87 / \underline{94.99}\\ 
    2:3         & 83.38     & 93.06 / \underline{93.22}& 94.92 / \underline{95.09}\\ 
    1:3         & 83.39     & 93.12 / \underline{93.31}& 94.99 / \underline{95.18}\\ 
    1:4         & \textbf{83.44}     & \textbf{93.14} / \textbf{\underline{93.41}}& \textbf{95.05} / \textbf{\underline{95.33}}\\ 
    1:5         & 82.72     & 92.19 / \underline{92.49}& 94.06 / \underline{94.36}\\  \bottomrule
    \end{tabular}}
    \label{tab:ab_pos_vs_neg_cub}
\end{table}

\subsection{More Ablation Studies}

The performance of our method w.r.t. different choices of the balancing hyper-parameter $\eta$ on the ImageNet-1K \cite{ILSVRC15} dataset are reported in Table~\ref{tab:ab_b_imagenet}. The trend is the same as that on the CUB-200-2011 \cite{WahCUB_200_2011} presented in the main body of our paper.
The best overall performance is achieved at the value of $2^{-5}=0.03125$ for $\eta$.

We discuss the effect of choosing different confidence thresholds $\tau_{1}$ and $\tau_{2}$, which are utilized for inducing the modulating factor. To include more samples for training, we enforce $\tau_{1}=\tau_{2}$, and denote them using the same $\tau$. The results are reported in Table~\ref{tab:ab_thres_cub} and Table~\ref{tab:ab_thres_cub_detr}. From the results, we can see that the binary-class detector (BCD) can be improved in a wide range of $\tau$, and our method is insensitive to this hyper-parameter. This indicates that as long as the easy samples are down-weighted, our weighted entropy (WE) loss is able to largely elevate the performance, and there is no need for carefully chosen hyper-parameters.

To further verify the effectiveness of our proposed WE loss, we compare the performance of adopting the standard entropy loss and our WE loss in the BCD. As reported in Table~\ref{tab:ab_entropy}, adopting the standard entropy loss can also improve the performance, and adopting our WE loss can always achieve better performance due to the ability to deal with the foreground-background imbalance in BCD.
On the CUB-200-2011 \cite{WahCUB_200_2011}, our WE loss surpasses the standard entropy loss by $0.41\%$, $0.53\%$, and $0.59\%$ in terms of \textit{Top-1 Loc}, \textit{Top-5 Loc}, and \textit{GT-known Loc} accuracies, respectively, when using the Top-1 localization results for evaluation.

Additionally, we tune the implicit foreground-background balancing factor in some detectors, e.g., the sampling ratio of positive to negative samples of the Faster R-CNN \cite{fasterrcnn15}. Results reported in Table~\ref{tab:ab_pos_vs_neg_cub} shows that our proposed WE loss can enable this kind of detectors to use more negative samples for training and learn more discriminative features.

\subsection{Visualization Comparison}

\begin{figure}[!t]
  \vspace{0.05in}
  \centering
  \includegraphics[width=2.8in]{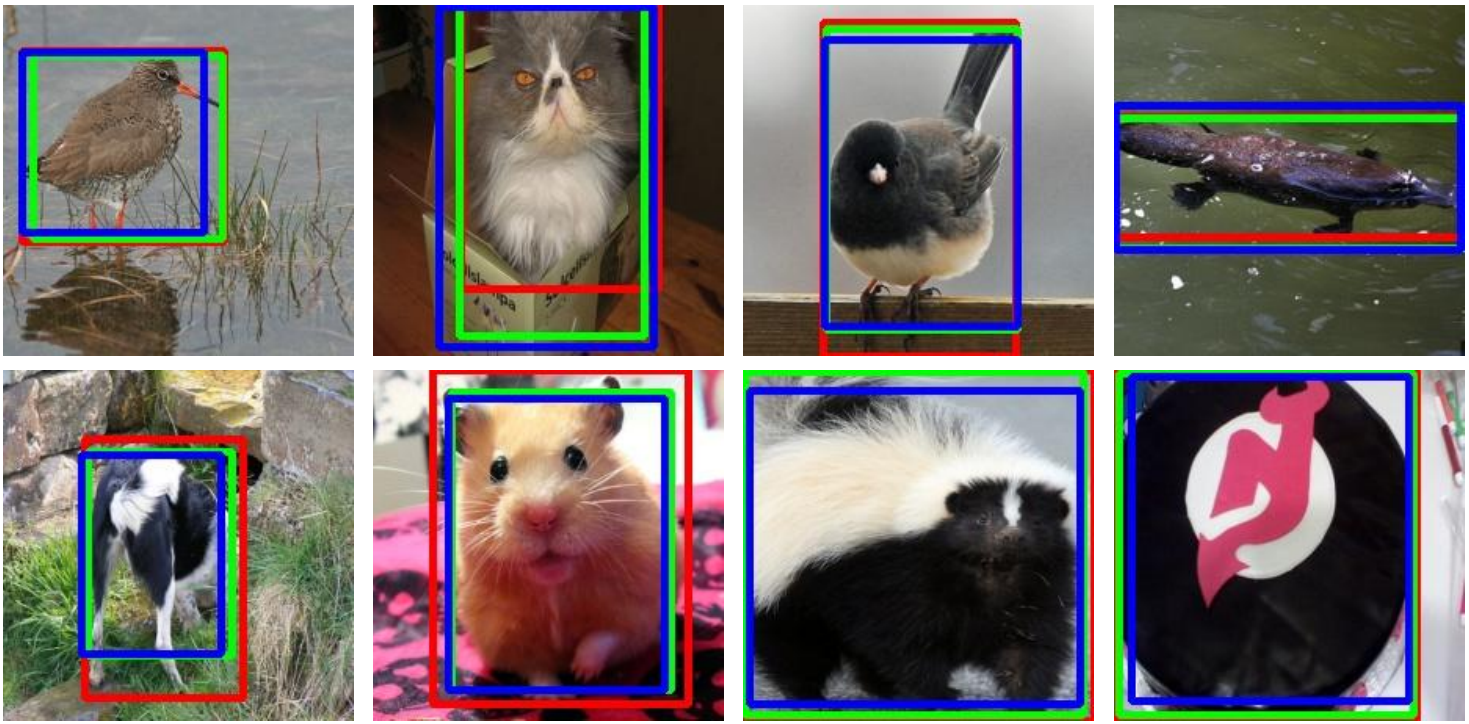}
  \caption{Visual comparison of our WEND and C$^{2}$AM \cite{C2AM22} on the ImageNet-1K (DenseNet161-Faster R-CNN). The red, blue, and green boxes denote the ground-truth bounding boxes, boxes predicted by C$^{2}$AM \cite{C2AM22}, and boxes predicted by our method, respectively.}
  \label{fig:visual_app}
\end{figure}

We provide more visual results on the ImageNet-1K \cite{ILSVRC15} in Figure~\ref{fig:visual_app}. These results demonstrate that our WEND can achieve more accurate localization than C$^2$AM \cite{C2AM22} in a wide variety of object categories.

\begin{figure}[!t]
  \vspace{0.05in}
  \centering
  \includegraphics[width=2.8in]{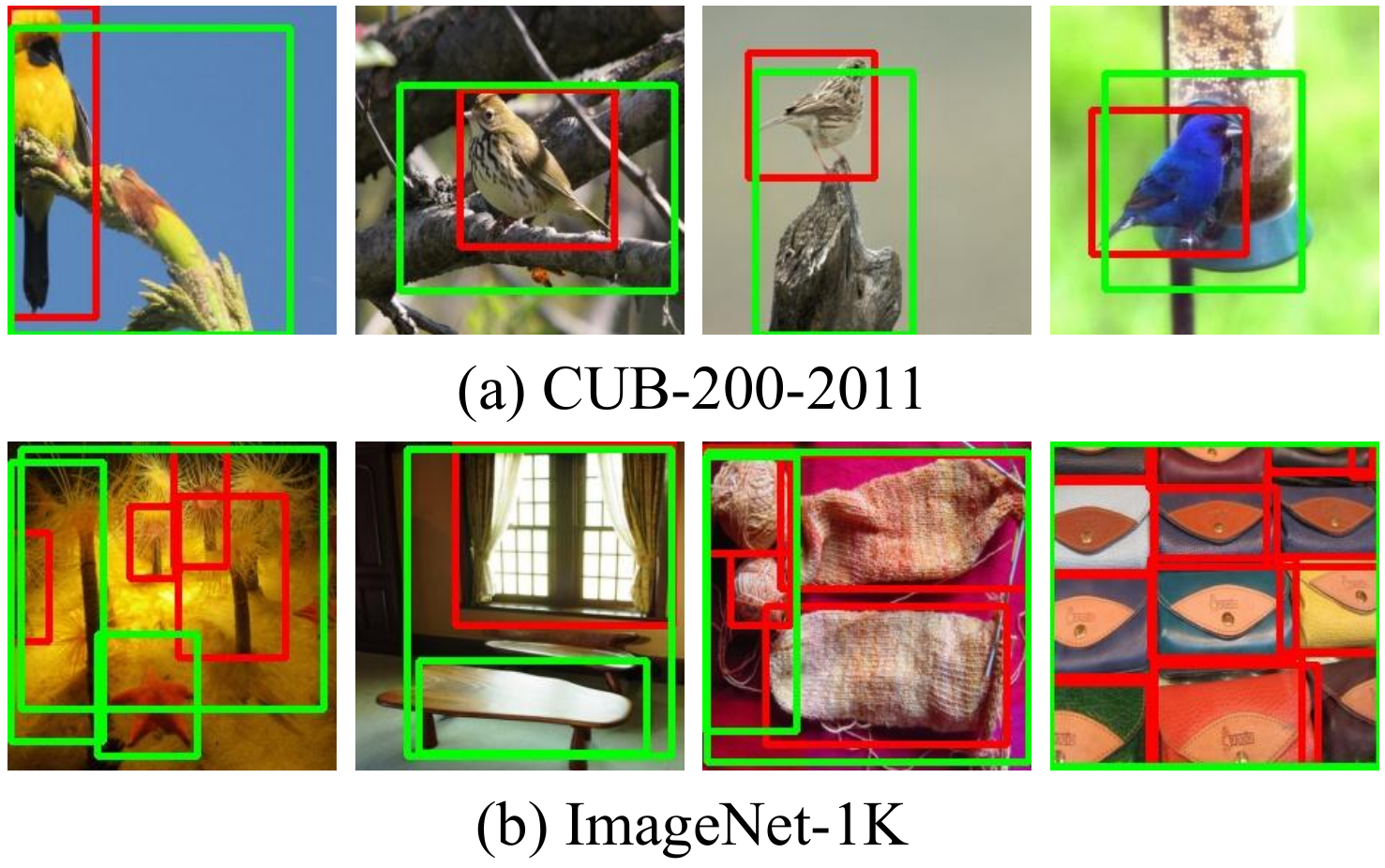}
  \caption{Failure cases of our WEND (DenseNet161-Faster R-CNN).}
  \label{fig:visual_failure}
\end{figure}

\section{More Discussions}

We also visualize some failure cases of our WEND to explore the focus in future work. As shown in Figure~\ref{fig:visual_failure}, on the CUB-200-2011 \cite{WahCUB_200_2011}, our WEND fails when the backgrounds are confusing. Thus a future work may be to discover the theme of the images to reduce distractions. It is the same for the more complicated dataset ImageNet-1K \cite{ILSVRC15}: when there exist more than one category of objects or confusing backgrounds, the algorithm has to be capable of finding the main target. In addition, the occlusion problem needs to be addressed.

\clearpage
\end{document}